%% file: main_arxiv.tex
\documentclass[runningheads]{llncs}

\usepackage[mobile]{eccv}

\newcommand{\sepappendix}{0}

\input{config.tex}

\usepackage[subtle]{savetrees} %

\usepackage{hyperref}

\usepackage{orcidlink}

\begin{document}

\title{Pano3D: Unified 3D Reconstruction and \\ Panoptic Segmentation}

\titlerunning{Pano3D: Unified 3D Reconstruction and Panoptic Segmentation}

\input{authors.tex}

\authorrunning{V. Barberteguy et al.}

\maketitle

\input{sections/00_abstract}
\input{sections/01_intro}
\input{sections/02_related_work}

\input{sections/03_method}
\input{sections/04_experiments}

\input{sections/05_ablations}

\input{sections/06_conclusion}

\input{sections/07_acknowledgements}

\bibliographystyle{splncs04}
\bibliography{main}

\clearpage
\section*{APPENDIX}
\input{supp_mat.tex}

\end{document}

%% file: config.tex
\usepackage{eccvabbrv}

\usepackage{graphicx}
\usepackage{booktabs}
\usepackage{amssymb}
\usepackage{mathtools}
\usepackage{colortbl}
\usepackage{pifont} %
\usepackage{multirow}

\usepackage[colorinlistoftodos, prependcaption, textsize=tiny]{todonotes}

\newcommand{\beginsupplement}{%
        \setcounter{table}{0}
        \renewcommand{\thetable}{A\arabic{table}}%
        \setcounter{figure}{0}
        \renewcommand{\thefigure}{A\arabic{figure}}%
        \setcounter{section}{0}
        \renewcommand{\thesection}{\Alph{section}}%
    }

\newcommand{\appendixref}[2]{%
    \ifnum\sepappendix=1 %
        #1%
    \else%
        #2%
    \fi%
}

\usepackage[accsupp]{axessibility}  %

%% file: authors.tex
\author{Victor Barberteguy\inst{1,2} \and
Ahmet Iscen\inst{1} \and 
Mathilde Caron\inst{1} \and
 \\ Alireza Fathi\inst{1} \and
 G\"{u}l Varol\inst{2}
\and Cordelia Schmid\inst{1}
}

\institute{
    Google DeepMind \and
    LIGM, \'Ecole des Ponts, IP Paris, Univ Gustave Eiffel, CNRS \\
    \email{victorbbt@google.com} \\
    {\tt\small \url{https://victorbbt.github.io/Pano3D}}
}

%% file: sections/00_abstract.tex
\begin{abstract}
Recent advances in 3D feedforward reconstruction neural networks have achieved remarkable success in dense reconstruction from images without any camera parameters. Yet, equipping  these models with robust semantic understanding remains an open problem. Here we introduce an approach that performs 3D reconstruction and 3D panoptic segmentation in a unified framework. We build on existing 3D reconstruction models and augment them with a set-based mask decoder. The approach is jointly trained with a geometric and semantic loss, which are shown to be mutually beneficial. More precisely, the features are initialized from the geometric information and then finetuned to capture jointly geometry and semantics. We demonstrate the generality of our approach by successfully applying our framework both to online and all-to-all attention reconstruction backbones.
Our method achieves state-of-the-art performance in 3D panoptic segmentation across ScanNet, ScanNet200, and ScanNet++ datasets. Ablation studies show that such joint training of a unified model equips 3D feedforward reconstruction neural networks with panoptic segmentation and yields mutually beneficial improvements.

\keywords{3D Panoptic Segmentation \and Scene Reconstruction}
\end{abstract}

\begin{figure}[ht]
    \centering
    \includegraphics[width=1.0\textwidth]{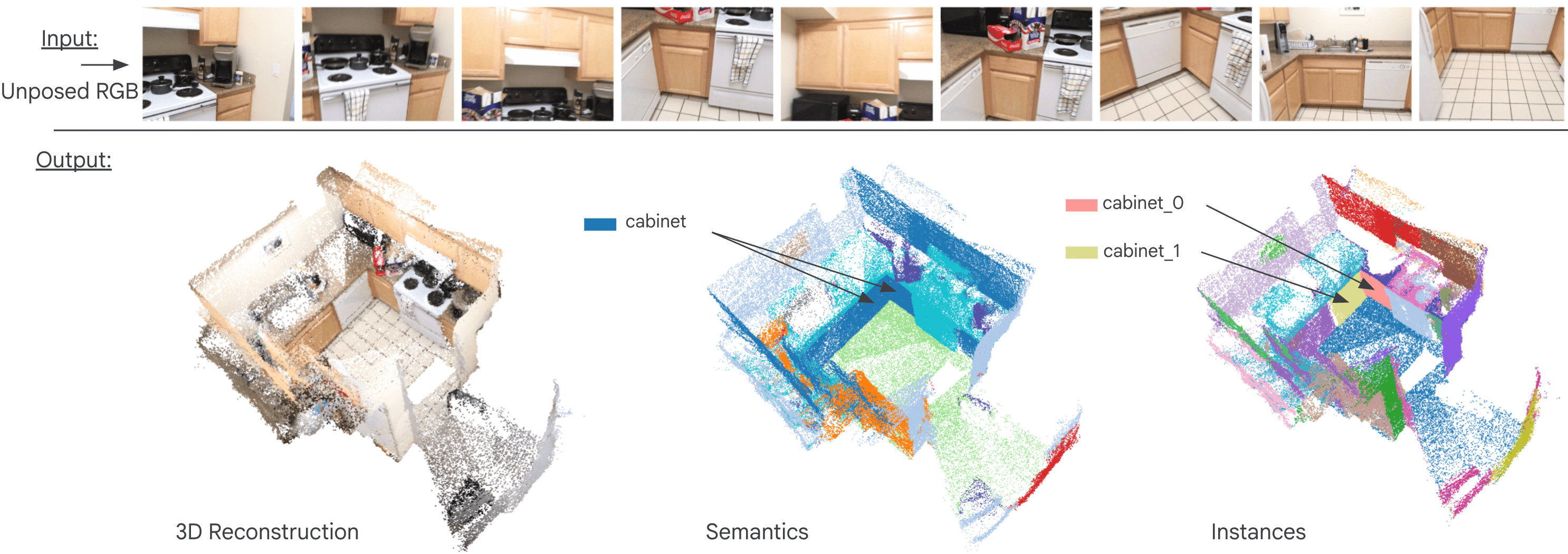}
    \caption{\textbf{Pano3D inputs and outputs:} Our model reconstructs a 3D point cloud (left) from unposed RGB frames, and predicts semantic classes (middle)  as well as semantic instance masks (right). Pano3D can jointly reconstruct a scene as well as detect instances and their classes.}
    \label{fig:kitchen_combined}
    \vspace{-.3cm}
\end{figure}

%% file: sections/01_intro.tex
\section{Introduction}
\label{sec:intro}

Holistic 3D scene understanding requires jointly reasoning about ``where'' things are (geometry) and ``what'' they are (semantics). Recent feedforward reconstruction models (FRM), such as DUSt3R~\cite{wang2024dust3r}, MASt3R~\cite{leroy2024grounding}, MUSt3R~\cite{cabon2025must3r}, and VGGT~\cite{wang2025vggt}, have demonstrated that transformers can reconstruct dense 3D pointmaps from unposed RGB images, opening new opportunities for unified 3D scene understanding. Several approaches~\cite{koch2025unified,fu2022panoptic,Hu2025PE3RP3,Sun2025Uni3RU3,Xu2025SIU3RSS} have started adding semantic capabilities to FRMs, either by freezing the backbone and relying on external 2D features~\cite{zust2025panst3r,Xu2025SIU3RSS,oquab2023dinov2}, or by adding dense contrastive heads that require unsupervised clustering at inference~\cite{koch2025unified,li_iggt:_2025,ranftl2021vision_dpt,McInnes_HDBScan_2017}.

In this paper, we introduce
a unified reconstruction and segmentation approach,
\textbf{Pano3D}, %
that eliminates the heuristic clustering algorithms in favor of a natively differentiable, query-based panoptic head appended to a multi-view reconstruction backbone.
Our approach is built on the simple insight that recognizing objects is implicitly encoded within feedforward reconstruction models. Indeed, these models contain 2D, monocular information in the encoder stage, and iteratively refine the representation to detect spatial boundaries and 3D structural coherence in the cross-view decoder, which, once combined, form a strong representation of the objects.
This makes Pano3D alignment-free: we directly predict instance-aware masks and class logits from learned queries in a feedforward manner, without relying on external 2D models or postprocessing. 

\begin{figure}[t] %
    \centering
    \includegraphics[width=1.0\textwidth]{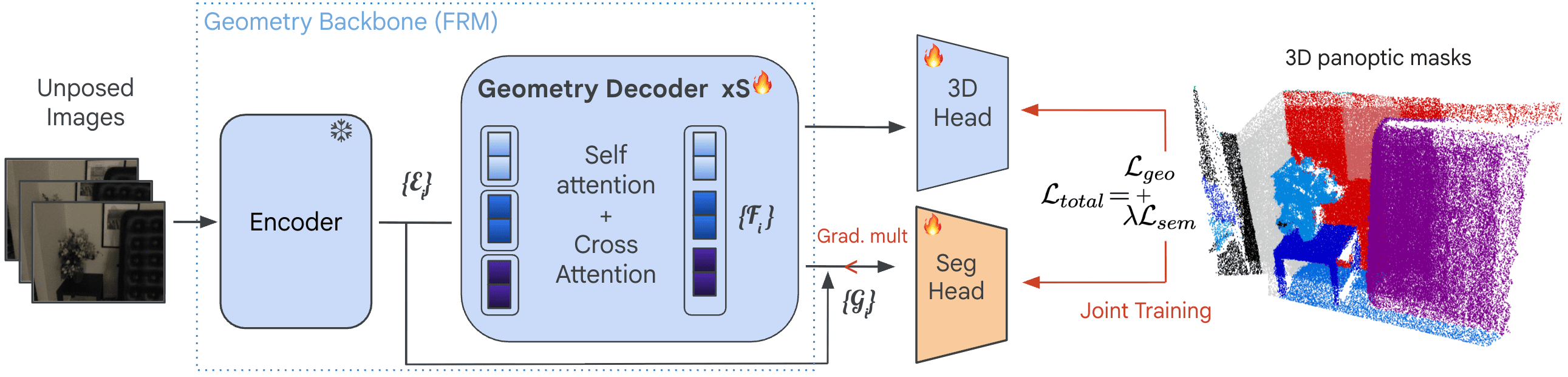}
    \vspace{-.3cm}
    \caption{\textbf{Framework overview:} Our method appends a segmentation head inspired from Mask2Former~\cite{cheng2022masked} on top of a feedforward reconstruction model. We extract features from the monocular encoder and the multi-view decoder of the FRM to predict 3D points as well as consistent masks across the input sequence. We jointly optimize the geometry decoder to bake in object understanding.}
    \label{fig:overview}
    \vspace{-.3cm}
\end{figure}

Our approach appends a set-based mask decoder directly to the output of the FRM. The decoder is similar to Mask2Former~\cite{cheng2022masked} in the 2D image segmentation field. By simply fusing the features from the encoder with the final cross-view decoder features 
we build features that are both precise and scene-aware, allowing the queries of the mask decoder to predict consistent masks across the entire input sequence.  Rather than freezing the network or crafting explicit consistency losses, we finetune the multi-view decoder on both 3D reconstruction and panoptic segmentation jointly. We demonstrate that this setting gives increased instance and semantic segmentation results while maintaining the reconstruction capability. Fig.~\ref{fig:kitchen_combined} illustrates the outputs from our model on an example input sequence: 3D reconstruction on the left, 3D semantic segmentation in the middle, and 3D semantic instance segmentation on the right.
Our core contributions are the following: \\
    $\bullet$~\textbf{Unified model:} We tackle 3D reconstruction and panoptic segmentation from unposed RGB views relying only on pretrained FRMs (see Fig.~\ref{fig:overview}). %
    Without relying on auxiliary 2D models and heuristic post-processing, Pano3D achieves state-of-the-art results on ScanNet~\cite{dai2017scannet}, ScanNet200~\cite{rozenberszki2022language_scannet200} and ScanNet++~\cite{yeshwanth2023scannet++} purely through feature adaptation. \\
    $\bullet$~\textbf{Backbone generality:} We apply our method to two different types of FRMs, namely MUSt3R~\cite{cabon2025must3r} and Pi3~\cite{wang2025pi}, demonstrating the %
    	adaptability of our method across online reconstruction models~\cite{cabon2025must3r} and all-to-all attention models\cite{wang2025pi}. \\
    $\bullet$~\textbf{Joint training:}
      We demonstrate that finetuning the geometry decoder, rather than freezing it, improves segmentation without compromising geometric fidelity. Furthermore, we highlight the importance of gradient scaling in balancing these two objectives.

%% file: sections/02_related_work.tex
\section{Related Work}
\label{sec:related}

\textbf{2D Panoptic Segmentation.} 
Panoptic segmentation~\cite{Kirillov2018PanopticS} consists in predicting a dense label map with instances and semantic labels for countable objects (`thing' classes: chair, water bottle, ...), and semantic labels only for `stuff' classes (sky, floor, ...). Early approaches ~\cite{he2017mask,cheng2020panoptic} extended CNN architectures by adding specialized heads. More recent transformer-based~\cite{dosovitskiy2020image} set-prediction methods, notably Mask2Former~\cite{cheng2022masked},
use bipartite matching over learnable object queries~\cite{carion2020end} to jointly predict thing and stuff classes. While successful in 2D, these methods lack spatial consistency across multiple views.

\noindent\textbf{Feedforward Reconstruction Models.} 
Recent models such as DUSt3R~\cite{wang2024dust3r} and MASt3R~\cite{leroy2024grounding} have redefined 3D reconstruction as a dense pointmap regression problem for pairs of unposed views. This paradigm naturally led to multi-view extensions: VGGT~\cite{wang2025vggt} and Pi3~\cite{wang2025pi} use all-to-all cross-attention for accurate but computationally expensive reconstruction, while MUSt3R~\cite{cabon2025must3r} introduces a memory mechanism for online, causal prediction. Our work builds upon these geometric backbones, building on the insight that their progressive refinement from monocular to multi-view features naturally encodes spatial discontinuities suitable for learning object instances.

\noindent\textbf{Unified 3D Semantic Transformers.}

Prior 3D panoptic methods operate on explicit representations such as raw point clouds from RGBD sensors as in ODIN~\cite{Jain2024ODINAS} and DeepViewAgg~\cite{robert2022learning_deepviewagg}, or postprocessed meshes as in OpenScene~\cite{peng2023openscene}, while others rely on implicit representations (NeRF~\cite{mildenhall2021nerf}, Gaussians~\cite{Kerbl20233DGS}) where semantic priors are learnt or distilled~\cite{zhou2024feature,Kobayashi2022DecomposingNF,fu2022panoptic}.
Building on FRMs, a new generation of feedforward models unifies reconstruction and semantics, often relying on CLIP features~\cite{radford2021learning} lifted in 3D~\cite{gong2025ov3r,Hu2025PE3RP3} or augmented with Gaussian heads~\cite{fan2024large_lsm,Sun2025Uni3RU3}. UNITE~\cite{koch2025unified} adds dense DPT heads~\cite{ranftl2021vision_dpt} on VGGT~\cite{wang2025vggt} but depends on knowledge distillation and explicit consistency losses. Both IGGT~\cite{li_iggt:_2025} (which uses SAM2~\cite{Ravi2024SAM2S} features) and UNITE require unsupervised clustering (HDBSCAN~\cite{McInnes_HDBScan_2017}) at inference, often yielding coarse masks.
PanSt3R~\cite{zust2025panst3r} places a Mask2Former-style~\cite{cheng2020panoptic} decoder on a frozen MUSt3R backbone with DINOv2~\cite{oquab2023dinov2} features and QUBO post-processing, but lacks an integrated framework for joint reconstruction and segmentation. SIU3R~\cite{Xu2025SIU3RSS} branches geometry and semantics early, requiring hand-crafted regularizations to enable inter-task benefits~\cite{chen2025sab3r}.
\textbf{Pano3D} builds on these works to design  a unified framework trained jointly with geometric and semantic losses. By employing a set-based query decoder on top of the geometry network, we avoid contrastive clustering and demonstrate that explicit consistency losses and heuristic post-processing are unnecessary. Our approach is generic and can be applied to various geometry backbones.

%% file: sections/03_method.tex
\begin{figure} %
    \centering
    \includegraphics[width=1.0\textwidth]{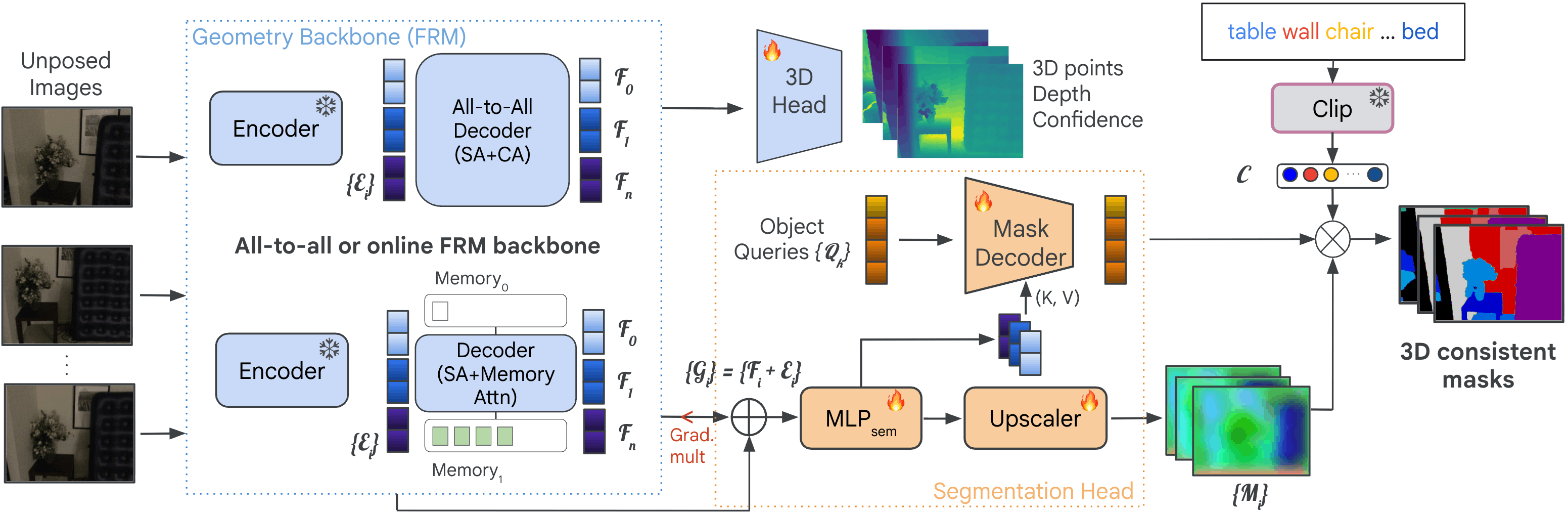}
    \caption{\textbf{Model architecture:} %
    	For a given input sequence $\mathcal{I} = \{I_i\}$, we extract features from the geometry encoder $\mathcal{E}_i$ and decoder $\mathcal{F}_i$ to create frame features $\mathcal{G}_i$ and mask features $\mathcal{M}_i$. The blue components (encoder, geometry blocks, 3D head) are from the geometry backbone (either it uses a memory-based architecture~\cite{cabon2025must3r} with causal memory attention %
    		or all-to-all attention~\cite{wang2025pi}). Learnable object queries $\mathcal{Q}_k$ are refined in the mask decoder by attending to $\mathcal{G}_i$. Then, they perform dot product against mask features $\mathcal{M}_i$ and 
    a matrix of frozen text embeddings $\mathcal{C}$ to predict semantic instance masks. Crucially, one object query predicts a set of 3D-consistent masks on the full input sequence. $\mathcal{F}_i$ are at the same time used to predict 3D information (pointmaps in reference frame's coordinate system, local pointmaps and confidence) with a linear head. As masks and pointmaps are pixel-aligned, we lift these masks to obtain 3D instance segmentation. We train the cross-view decoder
    on both the geometry and segmentation tasks. 
    }
    \label{fig:model}
    \vspace{-.3cm}
\end{figure}
\section{Methodology}
\label{sec:method}

We introduce Pano3D, a unified transformer framework for 3D reconstruction and segmentation, see Fig.~\ref{fig:model}. Given an unordered collection of $N$ unposed RGB images $\mathcal{I} = \{I_1, \dots, I_N\}$, our goal is to jointly recover the dense scene geometry (Sec.~\ref{sec:3Dreconstruction}) and a set of multi-view consistent 3D semantic instance masks (Sec.~\ref{sec:3DPanoptic_seg}) with joint finetuning of the geometry decoder (Sec.~\ref{sec:joint_opti}). We instantiate our framework by building on MUSt3R~\cite{cabon2025must3r} (memory-based online FRM in Fig.~\ref{fig:model}) or on Pi3~\cite{wang2025pi} (all-to-all FRM in Fig.~\ref{fig:model}) to extract features at the encoder and decoder stages (Sec.~\ref{sec:backbone_adaptation}).
The features from this encoder-decoder setup are then trained to both regress 3D information (3D head in Fig.~\ref{fig:model}) and to learn consistent object queries with a mask decoder similar to~\cite{Jain2024ODINAS}. The predicted masks and pointmaps are pixel-aligned (i.e., direct correspondence between a 3D predicted point and the masks), which makes the lifting to 3D straightforward.

\subsection{3D Reconstruction}
\label{sec:3Dreconstruction}

Our unified framework is built on top of a feedforward reconstruction model (FRM), which we briefly introduce in this section. Given a collection of $N$ unposed RGB images $\mathcal{I}=\{I_1, \dots, I_N\}$, the objective of the geometric backbone is to predict a dense pointmap $X_i \in \mathbb{R}^{H \times W \times 3}$ and an associated confidence map $C_i \in [0, \infty)$ for each frame $I_i$ in a globally aligned coordinate system. For an input image, patch features $E_i$ are first extracted using a pretrained monocular 2D encoder (e.g., CrocoV2~\cite{weinzaepfel2023croco} or DINOv2~\cite{oquab2023dinov2}). Following the training strategy detailed in Sec.~\ref{sec:exp:implem_details}, we keep this encoder frozen as we observe in our ablations (Sec.~\ref{sec:exp:ablations}) that finetuning end-to-end has a minor impact while increasing the training cost. These tokens are subsequently fed into a multi-view transformer decoder $\Phi_{\text{geo}}$, which aggregates information across viewpoints to output cross-view consistent geometric latent features $F_i$:
\begin{equation}
F_i = \Phi_{\text{geo}}(E_i, \mathcal{C}_{\text{context}}),
\end{equation}
\noindent where $\mathcal{C}_{\text{context}}$ represents the multi-view scene context. This setting was first introduced by DUSt3R~\cite{wang2024dust3r} for pairs of images. This context can be established either via an all-to-all cross-attention mechanism across all frames simultaneously (e.g., Pi3~\cite{wang2025pi}), 
or through a sequential memory tracking mechanism for online, streaming prediction (e.g., MUSt3R~\cite{cabon2025must3r}). 

A 3D prediction head then maps these refined latents to the spatial outputs: $\{X_i, C_i\} = \text{Head}_{\text{3D}}(F_i)$. The geometric backbone is optimized using a dense regression objective $\mathcal{L}_{\text{geo}}$:
\begin{equation}
\mathcal{L}_{\text{geo}} = \sum_{i=1}^{N} \sum_{p \in \Omega} C_{i,p} ||X_{i,p} - \hat{X}_{i,p}||_2 - \alpha \log(C_{i,p}),
\end{equation}
\noindent where $\Omega$ is the pixel grid and $\hat{X}_{i,p}$ denotes the ground-truth coordinates. The exact form of $\mathcal{L}_{\text{geo}}$ depends on the FRM (e.g., relative cameras and focal depth~\cite{wang2025pi}, or pointmaps in a global reference frame~\cite{cabon2025must3r}); we detail each instantiation in \appendixref{Sec.~A of the supplementary}{Appendix~\ref{sec:supp:implem}}.
As opposed to most methods integrating semantic understanding into FRMs~\cite{zust2025panst3r, koch2025unified}, our framework keeps $\mathcal{L}_{\text{geo}}$ activated and uses it as a geometric objective within our joint finetuning formulation (Eq.~\ref{eq:total_loss}) to prevent corrupting the reconstruction accuracy while optimizing for segmentation.

\subsection{3D Panoptic Segmentation}
\label{sec:3DPanoptic_seg}

We hypothesize that the high-dimensional feature space of a geometric backbone trained to align unposed images across  viewpoint changes already contains rich structural descriptors. Starting from monocular context, the cross-view decoder progressively recovers geometry and scene appearance~\cite{Stary2025UnderstandingMT}. 
Our approach thus extracts features at different stages of the underlying FRM, resulting in a unified architecture that does not need additional external models.

\noindent \textbf{Geometric Feature Bridge.} We introduce a lightweight adaptation module (cf.~Fig.~\ref{fig:model}, MLP$_{sem}$). For each frame $i$, we concatenate the features from the frozen encoder $E_i$ with the multi-view consistent features $F_i$ from the finetuned decoder. These are projected through an MLP adapter to produce patch features $\mathcal{G}_i$ for masked cross-attention with the object queries. The encoder, often trained in a self-supervised manner, contains local monocular context whereas the decoder brings scene-level consistency and geometry appearance.

\noindent \textbf{Set Transformers for Panoptic Segmentation.}  We adopt the set-prediction framework of Mask2Former~\cite{cheng2022masked} and extend it to sequences as proposed in ODIN~\cite{Jain2024ODINAS}. We initialize $K$ learnable object queries $\mathcal{Q} = \{q_k\}_{k=1}^K$ that act as sequence-level detectors; a single $q_k$ tracks a unique 3D instance across all $N$ images. For class predictions, we compute the similarity between query embeddings and a matrix of frozen CLIP~\cite{radford2021learning} text embeddings $\mathcal{C}$ (cf. Fig. \ref{fig:model}, right) as in~\cite{ghiasi2022scaling,Jain2024ODINAS,zust2025panst3r}. The semantic loss $\mathcal{L}_{sem}$ follows standard bipartite matching~\cite{cheng2022masked}. This allows Pano3D to be queried on any text at inference time, simply by substituting the text matrix with new embeddings. We provide the loss specifics in  \appendixref{Sec.~A of the supplementary material}{Appendix~\ref{sec:supp:implem}}.

\noindent \textbf{High-Resolution Mask Features.} To recover sharp boundaries, we upscale the adapted patch features by a factor of 4 (for patch size 16, it results in one-quarter of the input resolution) 
using sequential pixel-shuffle operations in the upscaler $U_{sem}$, resulting in the dense mask features $\mathcal{M}_i$. Because our queries $\mathcal{Q}$ attend to the combination of these features, they are provided with both local appearance and cross-view (scene level) context, enabling consistent 3D segmentation.

Crucially, we do not stop the gradients on the decoder features $F_i$ in the concatenation, providing a direct gradient path from $\mathcal{L}_{sem}$ into the geometry decoder through both the patch features $\mathcal{G}_i$ and the mask features $\mathcal{M}_i$ (Sec.~\ref{sec:joint_opti}). Following the Mask2Former method~\cite{cheng2022masked}, we supervise the object queries at each layer of the mask decoder to have more stable and faster convergence.

\subsection{Joint Training}
\label{sec:joint_opti}
Existing work either freeze the FRM~\cite{zust2025panst3r} or add a pixel-wise consistency loss to  the multi-task learning~\cite{koch2025unified} to prevent corrupting the geometric quality.
While FRMs learn precise local, low-level matching, they are explicitly trained to discard any ambiguous prediction in textureless areas or reflective surfaces~\cite{wang2024dust3r}. We identify that semantic instance segmentation can act as an implicit object prior that mitigates this limitation.

By unfreezing the geometric decoder $\Phi_{geo}$ and allowing the semantic gradients from $\mathcal{L}_{sem}$ to flow back from the bipartite matching loss into the geometric features, the network learns to group pixels into discrete entities.
Our combined objective is:
\begin{equation}
    \mathcal{L}_{total} = \mathcal{L}_{geo} + \lambda \mathcal{L}_{sem},
\label{eq:total_loss}
\end{equation}
where $\lambda$ balances the two tasks.
As we demonstrate in our experiments (Sec.~\ref{sec:experiments}), this joint optimization yields multi-view consistent masks without the need for heuristic post-processing algorithms such as Quadratic Unconstrained Binary Optimization (QUBO)~\cite{zust2025panst3r} and without enforcing any explicit prior with hand-crafted consistency losses. We refer to Sec.~\ref{sec:exp:ablations} for a study on loss balancing to show that geometry performance is maintained when training on both tasks.

\noindent \textbf{Segmentation Gradient Scaling.} Besides the loss balancing, we apply a gradient multiplier during backpropagation to part of the network. Specifically, when training, we first log the per-task gradients and observe that the 
semantic gradients are typically an order of magnitude higher than the geometric gradients, which could perturb the weights of the multi-view decoder excessively.
We therefore measure the scale between the norm of the two gradients when trained with no gradient multiplier, and apply this as
a constant scaling factor $\gamma = 0.1$ to the gradients originating from the segmentation head before they enter the geometric decoder $\Phi_{geo}$ during backpropagation (as denoted in Fig.~\ref{fig:model} with `Grad.\ mult.'). As demonstrated by our ablation studies in Sec.~\ref{sec:exp:ablations}, while this gradient scaling has a negligible impact on segmentation metrics, it serves as a safeguard for geometry, ensuring improved 3D consistency when jointly finetuning the geometry decoder.

\subsection{Adaptation to Different FRMs}
\label{sec:backbone_adaptation}
Our framework is flexible and can be plugged with memory-based reconstruction models such as %
MUSt3R~\cite{cabon2025must3r}, as well as all-to-all attention models such  as Pi3~\cite{wang2025pi}.

\noindent\textbf{Memory Models.} To handle large image collections beyond device memory limits, we leverage the recurrent memory mechanism of MUSt3R~\cite{cabon2025must3r}, but extend it to maintain semantic instance consistency in the queries. Our model is thus equipped with two distinct memories. The first is an object-centric memory of the scene (object queries), and the other one is the original frame memory of MUSt3R (containing the patch tokens). As we finetune the model, the frame tokens learn some object priors on top of the original geometric cues inherited from the pretrained model. The processing happens in two phases. In Phase 1 (Update), the model processes keyframes to build the global geometric memory $C_{context} = \mathcal{M}em_{geo}$. These frame features $\mathcal{G}_i^{update}$ are used to initialize the semantic queries $\mathcal{Q}_k^{update}$ %
in the mask decoder $\Phi_{panoptic}$. In Phase 2 (Render), the same or new frames attend to $\mathcal{M}em_{geo}$ to
grasp the scene context. Then, the queries are updated against the new frames features $\mathcal{G}_i^{render}$ to produce $\mathcal{Q}_k^{render}$. A final dot product between $\mathcal{Q}_k^{render}$ and $\mathcal{M}_i^{render}$ gives the mask predictions:
\begin{equation}
    \mathcal{Q}_{k}^{render} = \Phi_{panoptic}(\mathcal{Q}_k^{update}, \mathcal{G}_i^{render}).
\end{equation}

As in MUSt3R~\cite{cabon2025must3r}, we supervise both phases during training, but the object queries need particular care. The queries are initialized in the update phase, and the render phase reuses the updated query state. 
Importantly, we concatenate the mask predictions from both phases along the view axis, and perform a single bipartite matching against the scene-level ground truth. This enforces that a given query $\mathcal{Q}_k$ must detect the same object instance in both the update and render frames, naturally yielding consistent assignments when filling the memory, or when rendering new frames. We provide a detailed figure and the specifics on the training of the memory in \appendixref{Sec.~A of the supplementary.}{Appendix~\ref{sec:supp:implem}.

\noindent\textbf{All-to-all Attention Models.} Equipping this type of FRMs~\cite{wang2025pi, wang2025vggt} with our panoptic decoder is more straightforward as we use all the frames as context. For a given input sequence, we extract the encoder features and decoder features to form the $\mathcal{G}_i$, that are fed %
as key and values for cross-attention with the object queries, and upscaled $\mathcal{M}_i$ for the final dot product with the updated $\mathcal{Q}_k$.

}

%% file: sections/04_experiments.tex
\section{Experiments}
\label{sec:experiments}

We first provide implementation details (\cref{sec:exp:implem_details}), followed by benchmark descriptions (\cref{sec:exp:benchmarks}). We then
compare against the state of the art (\cref{sec:exp:sota})
and ablate model components (\cref{sec:exp:ablations}).
Finally, we present qualitative results (\cref{sec:exp:qualitative}).

\subsection{Implementation Details}
\label{sec:exp:implem_details}
\textbf{Architecture.} We use the MUSt3R~\cite{cabon2025must3r} architecture as the online geometry decoder initialized with pretrained weights, and the Pi3 \cite{wang2025pi} pretrained model for all-to-all geometry model. The mask decoder is a 9-layer decoder  
 initialized from Mask2Former~\cite{cheng2022masked}. We use the decoder head trained on top of %
 ResNet50 features~\cite{he2016deep_resnet} on COCO~\cite{lin2014microsoft_coco} dataset.
  We use 100 queries for ScanNet~\cite{dai2017scannet} and ScanNet200~\cite{rozenberszki2022language_scannet200}, and 200 queries for ScanNet++~\cite{yeshwanth2023scannet++}.

\noindent\textbf{Training.} All models are trained on TPU v6 accelerators using the AdamW optimizer~\cite{kingma2017adammethodstochasticoptimization}. We use a base image resolution of $384 \times 512$ for MUSt3R~\cite{cabon2025must3r} and $378 \times 518$ for Pi3~\cite{wang2025pi}. For training we use $N=15$ views. Following standard practice, we apply random color jitter and random stride between 1 and 4 for frame selection.

\noindent\textbf{Loss Weights.} We use the Mask2Former~\cite{cheng2022masked} loss formulation with weights $\lambda_{cls}$, $\lambda_{dice}$ and $\lambda_{bce}$ set to 2, 2 and 5 following the original paper. The geometric reconstruction loss is weighted by a factor of 10 relative to the segmentation loss. We scale the segmentation gradients before backpropagation in the geometry decoder by a factor 0.1. %

\noindent\textbf{Inference.} Unless stated otherwise, we evaluate our online model in its \textit{render phase}. We select up to 20 frames via uniform linear spacing (linspace) to build the geometric memory and initialize object queries. We then perform a second render pass on all evaluation frames to generate dense, consistent predictions. For the all-to-all model, we pass all the frames into the model.

We refer to \appendixref{Sec. A of the supplementary material}{Appendix~\ref{sec:supp:implem}} for more details on the architecture and the training.

\subsection{Datasets and Benchmarks}
\label{sec:exp:benchmarks}

\noindent \textbf{Datasets.} We evaluate our method across three indoor datasets, ScanNet~\cite{dai2017scannet}, ScanNet200~\cite{rozenberszki2022language_scannet200} and ScanNet++~\cite{yeshwanth2023scannet++}. %
\textbf{ScanNet}~\cite{dai2017scannet} is a standard dataset for 3D scene understanding. It consists of 1,513 scans with 20 semantic categories. We subsample every 20th frame of the original sequences.
\textbf{ScanNet200}~\cite{rozenberszki2022language_scannet200} uses the same physical scenes as ScanNet but expands the label set to 200 categories. This introduces a challenging long-tail distribution that requires more fine-grained understanding. 
\textbf{ScanNet++}~\cite{yeshwanth2023scannet++} was captured with laser scanners and DSLR cameras that yield an improved depth resolution compared to the depth sensors used previously. It also contains dense annotations of more than 1000 different object types which are remapped and truncated to the top 100 classes for the standard semantic benchmark.

\noindent \textbf{Benchmark Protocols.} To
assess Pano3D, we adopt two 3D-centric evaluation protocols concurrently established by UNITE~\cite{koch2025unified} and IGGT~\cite{li_iggt:_2025}.\\
(1) \textit{Mesh-based Evaluation (UNITE):} The semantic performance is evaluated independent from geometric imprecision. 
 To do so, UNITE maps its semantic predictions on the GT clean mesh used in official benchmarks. They lift their predicted features to 3D, downsample and aggregate them in a local neighborhood before transferring them on the mesh with nearest neighbor. They report mean 3D intersection over union (3D mIoU) and mean accuracy (mAcc) for semantic segmentation, and class-agnostic average precision AP, AP25, and AP50 for 3D instance detection.\\
(2) \textit{Voxel-based Evaluation (IGGT):} IGGT evaluates the semantic predictions jointly with the predicted geometry. To do so they discretize the GT and predicted 3D point cloud into a voxel grid (voxel size $0.1$m). A prediction is a true positive only if the model correctly predicts both the spatial occupancy (reconstruction) and the semantic category (segmentation). We report the 3D mIoU on voxelized predictions alongside standard per view and Multi-View (MVD) with median alignment geometric metrics, including Absolute Relative Error (Abs. Rel) and Inlier Ratio ($\tau$) at a 1.03 threshold. These two complementary protocols enable benchmarking 3D semantic understanding and depth accuracy separately. This evaluation is more challenging than mesh aggregation, as nearest-neighbor interpolation on a mesh can smooth over small geometric drifts, whereas voxelization strictly penalizes spatial misalignment.

\subsection{Comparison with the State of the Art}
\label{sec:exp:sota}

To compare to the most closely related state-of-the-art methods, the concurrent work UNITE~\cite{koch2025unified} and the very recent work IGGT~\cite{li_iggt:_2025},  we use their respective evaluation protocols and report these comparisons in separate sections. 

\subsubsection{Comparison to the State of the Art with the UNITE Metrics~\cite{koch2025unified}.}

For both tasks in the UNITE benchmark, we first detail how we transfer our predictions on the mesh, and then compare to other RGB only methods.

\textbf{$\bullet$ Semantic Segmentation (Tab.~\ref{tab:semantic_results}).} To evaluate on the mesh, we aggregate the class logits from all the queries (marginalization) to get logits on the frames. We then lift the logits map using GT depth information and aggregate them on the mesh via k-nearest neighbors. Finally, we interpolate the class logits on the mesh and then take the most probable class at each 3D point.

\input{tables/semseg3d}

\input{tables/inst3d}

Tab.~\ref{tab:semantic_results} reports results for 3D semantic segmentation on ScanNet, ScanNet200 and ScanNet++. We observe that Pano3D achieves state-of-the-art results by a margin. It  outperforms UNITE by 16.6 mIoU on ScanNet, 10.1 mIoU on ScanNet200 and 12.5 mIoU on ScanNet++. While UNITE attempts to distill a large amount of teacher information (instances, semantics, articulations) via contrastive clustering, Pano3D directly optimizes for the target task using a set-based decoder, allowing it to detect instances more accurately. 
We note that the mIoU performance gap is larger  than that of mAcc. This can be explained by the fact that UNITE clusters high-dimensional features, which can cause the mask to bleed onto the background. Hence it covers the ground truth mask but results in more false positives, which reduce the mIoU but not the mAcc. 
In contrast, our mask decoder's Dice loss directly optimizes for IoU, resulting in crisper boundaries and fewer false positives.

Furthermore, Pano3D outperforms PanSt3R~\cite{zust2025panst3r} by an even larger margin. PanSt3R  relies on frozen geometric features combined with 2D DINOv2 features and selects a subset of "memory frames" for feature extraction. In contrast, Pano3D’s global object queries attend to the entire image sequence. This ensures full scene coverage and better agreement from multi-view information, whereas static frame selection can miss objects that appear in views which were not selected.

\textbf{$\bullet$ Instance Segmentation (Tab.~\ref{tab:instance_results})}. In Pano3D, each query predicts mask logits for each pixel of the entire sequence. Thus, it's likely that some queries predict masks logits for the same pixels (e.g., a mouse pad might get a mask proposal, but the query that detects the desk on which it sits might want to integrate the mouse pad into its mask). To resolve this, we aggregate the mask proposals on the mesh (weighted by k nearest neighbors with $k=5$), and use a winner-takes-all approach (the most confident query at each 3D point is kept) without relying on expensive postprocessing.

Tab.~\ref{tab:instance_results} evaluates class-agnostic 3D instance segmentation. It shows consistent performance gain of Pano3D compared to other works on ScanNet and ScanNet200, for example +9.3 in AP50 on ScanNet over the second best method UNITE.  Because our object queries are optimized concurrently with the geometric backbone, they natively learn to produce 3D-consistent masks. The superior AP scores on ScanNet, ScanNet200 and ScanNet++ demonstrate that our queries are more effective at delineating instances than clustering-based alternatives. On ScanNet++ that contains more instances per scene on average, and for which we train 200 queries, Pano3D struggles to detect accurately instances, leading to only slight improvement over UNITE in AP and AP50. While AP and AP50 rewards detection with accurate boundaries, AP25 rewards detecting additional small objects with less precision. This is possible with the clustering approach used by UNITE, and is the reason why our scores are higher in AP, but lower in AP25. If we detect an instance, we detect it more precisely than instance feature learning frameworks that need clustering. However, as the number of instances gets larger in the scene, Pano3D can miss some objects that UNITE might detect by finding small clusters in 3D. We also observe that the model sometimes groups instances that are close into the same mask as observed in Fig.~\ref{fig:failure_clutter}. This harms object detection, but has less impact on semantics where we outperform other methods by a larger margin (Table~\ref{tab:semantic_results}). It is also worth highlighting that even though UNITE distills CLIP features whereas we train on the labels, this evaluation is \textbf{class-agnostic} and shows our masks are of superior quality, set aside the classes.

\input{tables/coverage_iggt}
\subsubsection{Comparison to the State of the Art with the IGGT Voxel-based Evaluation~\cite{li_iggt:_2025}.}
Following IGGT~\cite{li_iggt:_2025}, we evaluate the  3D consistency of our masks using a voxel-based metric on their subset of the ScanNet++ benchmark. It consists of 9 scenes from ScanNet++ containing each around 10 frames. 
We align our predictions with Procrustes alignment~\cite{LUOProcrustes} and apply density and confidence filter to remove outliers.

As observed in Tab.~\ref{tab:coverage_3d}, our MUSt3R model is geometrically less precise than the VGGT model~\cite{wang2025vggt} used by IGGT. Indeed, VGGT is a significantly larger model, i.e., VGGT contains more than 1B parameters, whereas MUSt3R  only has 468M parameters. This translates to a slightly low  3D reconstruction accuracy of our Pano3D model compared to IGGT. However, the semantic performance including the 3D mIoU is higher for our Pano3D model. We can also observe that learning instances and semantics in the model slightly degrades the per-view reconstruction performance. When using Pi3~\cite{wang2025pi} which is closer in size and quality to VGGT~\cite{wang2025vggt}, we notice an improvement in multi-view depth while maintaining the per-view reconstruction performance, and even better downstream segmentation than with MUSt3R~\cite{cabon2025must3r}. It suggests that better FRMs naturally produce stronger features for segmentation.
However, we acknowledge that IGGT is a fully open-vocabulary method since it's trained on a very diverse set of instance masks, and then queries frozen 2D models. Hence, our in-domain training naturally yields better segmentation, and the alignment and filtering pipeline reduces the impact of geometric errors. We also highlight that being an open-vocabulary method, their pipeline is more cumbersome and not feedforward only: they use a heavier model, use clustering to produce masks, reproject the masks in 2D and call external specialized models for semantics~\cite{ghiasi2022scaling}.

%% file: tables/semseg3d.tex
\begin{table} [t] %
\caption{\textbf{3D semantic segmentation.} Comparison on ScanNet, ScanNet200 (open-vocabulary), and ScanNet++ with other RGB-only methods using the UNITE metrics. We report 3D mIoU and mAcc.  $\dagger$ denotes evaluation by~\cite{koch2025unified}.
}
\vspace{-.3cm}
\label{tab:semantic_results}
\centering
\footnotesize
\begin{tabular}{l cc cc cc}
\toprule
& \multicolumn{2}{c}{ScanNet} & \multicolumn{2}{c}{ScanNet200} & \multicolumn{2}{c}{ScanNet++} \\
\cmidrule(lr){2-3} \cmidrule(lr){4-5} \cmidrule(lr){6-7}
Method & mIoU$_{\text{3D}}\uparrow$ & mAcc$_{\text{3D}}\uparrow$ & mIoU$_{\text{3D}}\uparrow$ & mAcc$_{\text{3D}}\uparrow$ & mIoU$_{\text{3D}}\uparrow$ & mAcc$_{\text{3D}}\uparrow$ \\
\midrule
Pe3R$^\dagger$~\cite{Hu2025PE3RP3} & 10.7 & 19.8 & \textcolor{white}{0}2.5 & \textcolor{white}{0}6.5 & \textcolor{white}{0}8.3 & 16.0 \\
Uni3R$^\dagger$~\cite{Sun2025Uni3RU3} & 29.3 & 39.4 & \textcolor{white}{0}4.1 & \textcolor{white}{0}6.8 & \textcolor{white}{0}5.2 & 10.8 \\
LSM$^\dagger$~\cite{fan2024large_lsm} & 32.2 & 41.1 & \textcolor{white}{0}6.3 & \textcolor{white}{0}9.7 & 14.3 & 26.5 \\
PanSt3R$^\dagger$~\cite{zust2025panst3r} & 42.6 & 49.7 & 13.3 & 19.9 & 21.6 & 31.4 \\
UNITE~\cite{koch2025unified} & 48.7 & 68.3 & 14.5 & 26.3 & 17.2 & 37.0 \\
\rowcolor[gray]{0.9} \textbf{Pano3D (MUSt3R)} & \textbf{65.3} & \textbf{76.6} & \textbf{24.6} & \textbf{32.9} & \textbf{29.7} & \textbf{42.0} \\
\bottomrule
\end{tabular}
\end{table}

%% file: tables/inst3d.tex
\begin{table}[t] %
\caption{\textbf{Class-agnostic 3D instance segmentation. } Similar to Tab.~\ref{tab:semantic_results}, we report the UNITE metrics. $\dagger$ denotes evaluation by~\cite{koch2025unified}.
}
\vspace{-.3cm}
\label{tab:instance_results}
\centering
\begin{tabular}{@{}l ccc ccc ccc@{}}
\toprule
& \multicolumn{3}{c}{ScanNet} & \multicolumn{3}{c}{ScanNet200} & \multicolumn{3}{c}{ScanNet++} \\
\cmidrule(lr){2-4} \cmidrule(lr){5-7} \cmidrule(lr){8-10}
Method & AP & AP$_{50}$ & AP$_{25}$ & AP & AP$_{50}$ & AP$_{25}$ & AP & AP$_{50}$ & AP$_{25}$ \\
\midrule
SAM3D~\cite{Yang2023SAM3DSA} & \textcolor{white}{0}6.3 & 17.9 & 47.3 & 12.1 & 28.6 & 54.1 & 3.0 & \textcolor{white}{0}7.9 & 22.3 \\
PanSt3R$^\dagger$~\cite{zust2025panst3r} & 11.4 & 29.3 & 53.4 & 10.6 & 27.3 & 50.8 & 6.5 & 15.9 & 33.9 \\
UNITE~\cite{koch2025unified} & 13.2 & 29.6 & 57.2 & 12.3 & 28.8 & 58.2 & 6.6 & 16.1 & \textbf{40.4} \\
\rowcolor[gray]{0.9} \textbf{Pano3D (MUSt3R)} & \textbf{15.7} & \textbf{38.9} & \textbf{66.9} & \textbf{14.6} & \textbf{36.8} & \textbf{59.7} & \textbf{6.9} & \textbf{18.2} & 39.7 \\
\bottomrule
\end{tabular}
\end{table}

%% file: tables/coverage_iggt.tex
\definecolor{highlight}{gray}{0.95}
\newcommand{\cmark}{\ding{51}}
\newcommand{\xmark}{\ding{55}}

\begin{table}[t]
\caption{\textbf{Main results on the IGGT benchmark (ScanNet++).} We evaluate Pano3D on joint reconstruction and semantic understanding following the protocol of IGGT~\cite{li_iggt:_2025}. We report geometric fidelity (AbsRel, $\tau$) per frame and multi-view (MVD), sequential tracking metrics, and semantic segmentation in both 2D and 3D. All 3D mIoU scores are evaluated on voxelized predictions. $^\ddagger$~denotes our reproduction of the model. $^\ast$~denotes our evaluation with the official inference script. The rest of IGGT scores and previous works are evaluated by IGGT~\cite{li_iggt:_2025}.
}
\vspace{-.3cm}
\label{tab:coverage_3d}
\centering
\resizebox{\textwidth}{!}{
\begin{tabular}{l | cc cc | cc}
\toprule
& \multicolumn{4}{c|}{\textbf{Reconstruction}} & \multicolumn{2}{c}{\textbf{Semantics}} \\
\textbf{Method} & AbsRel $\downarrow$ & $\tau$ $\uparrow$ & AbsRel (MVD) $\downarrow$ & $\tau$ (MVD) $\uparrow$ & 2D mIoU $\uparrow$ & 3D mIoU $\uparrow$ \\
\midrule
\multicolumn{7}{c}{\cellcolor{highlight} \textbf{ScanNet++}} \\
LSeg~\cite{li2022language_seg} & n/a & n/a & n/a & n/a & 22.61 & n/a \\
OpenSeg~\cite{ghiasi2022scaling} & n/a & n/a & n/a & n/a & 13.92 & n/a \\
Feature-3DGS~\cite{zhou2024feature} & 5.92 & 41.64 & -- & -- & 22.47 & 10.59 \\
LSM~\cite{fan2024large_lsm} (Multi-View) & 2.96 & 83.28 & -- & -- & 17.88 & 15.17 \\
VGGT~\cite{wang2025vggt} & 2.75 & 85.41 & -- & -- & -- & -- \\
IGGT~\cite{li_iggt:_2025} & 2.61 & 85.66 & \textcolor{white}{0}2.73$^\ast$ & \textcolor{white}{0}83.90$^\ast$ & 31.31 & 20.14 \\
MUSt3R$^\ddagger$~\cite{cabon2025must3r} & 2.95 & 81.87 & 2.88 & 80.62 & n/a & n/a \\
Pi3$^\ddagger$~\cite{wang2025pi} & \textbf{2.44} & \textbf{87.50} & 2.49 & 83.40 & n/a & n/a \\
\rowcolor[gray]{0.9} Pano3D (MUSt3R)     & 3.10 & 78.21 & 2.59 & 74.81 & 46.61 & 24.88 \\
\rowcolor[gray]{0.9} Pano3D (Pi3) & \textbf{2.44} & 87.30 & \textbf{2.42} & \textbf{84.93} & \textbf{52.54} & \textbf{29.46} \\
\bottomrule
\end{tabular}
}
\vspace{-.3cm}
\end{table}

%% file: sections/05_ablations.tex
\subsection{Ablation Studies}
\label{sec:exp:ablations}
We validate our core architectural and training choices through a series of ablation studies on the ScanNet validation set. 
We evaluate 3D mIoU and class agnostic 3D AP following UNITE~\cite{koch2025unified} as detailed in Sec.~\ref{sec:exp:benchmarks}. For 2D mIoU and 2D mAcc, as well as multi-view depth estimation (MVD), we take 20 frames from the validation set (we first uniformly sample 200 frames, and select the first 20). 
In MVD, we apply standard median alignment on the entire sequence whereas the per view depth evaluated in Tab.~\ref{tab:coverage_3d} scales each frame independently. Unless specified otherwise, we use the online version of our method for ablations, building on the MUSt3R reconstruction model~\cite{cabon2025must3r}.

\noindent \textbf{Joint Finetuning.}
A central claim of our methodology (Sec.~\ref{sec:joint_opti}) is the introduction of a unified model trained jointly on both 3D reconstruction and panoptic segmentation. Tab.~\ref{tab:ablation} (1) shows how joint finetuning benefits both tasks: while finetuning the model only on semantics (\textit{FT Semantics Only}) causes the geometric latent space to be completely corrupted—evidenced by the catastrophic increase in AbsRel—our joint optimization (\textit{G + S})
preserves the geometric fidelity of the original FRM. A critical insight is that training on geometry and semantics increases the MVD performance compared to the frozen model, as also evidenced in Tab.~\ref{tab:coverage_3d}. Even though the per frame depth metrics are slightly degraded (meaning the predictions are a little blurred), the geometric layout at the scene level is better (less drift or unconfident predictions). However, the MVD final performance is still below MUSt3R~\cite{cabon2025must3r} finetuned only on geometry.

It is also worth noting that joint training provides a more important boost in instance localization rather than semantics, increasing AP by 4.9 compared to the frozen baseline (15.7 vs 10.8). This suggests that the joint optimization does not merely inject semantic labels, but actively refines the model's ability to delineate 3D object boundaries and maintain multi-view consistency. This confirms that semantic gradients act as an implicit structural regularizer. Finally, unfreezing the encoder and finetuning the full architecture end-to-end yields improvement in geometry, but loses semantic capacity over Pano3D, while increasing the training cost.

\input{tables/ablations}

\noindent \textbf{Combining Encoder and Decoder Features.}
We investigate the necessity of our multi-level feature adaptation. To do so, we provide the mask decoder head only with encoder or decoder features. As shown in Tab.~\ref{tab:ablation} (2), using only the initial encoder features yields poorer performance. The mask decoder can infer multi-view masks by cross attending to the full sequence, but has \textbf{-3} mIoU as it lacks multi-view consistency of the cross-view decoder. Using only the decoder features yields a closer performance in segmentation, yet the depth metrics are lower compared to the final version, as the decoder loses some capacity to refine geometry. This confirms that object representations are distributed across the hierarchy of the geometric backbone, from local monocular context in the encoder to cross-view scene aware features, and that a simple skip connection in Pano3D provides both capacity and efficiency to reconstruct and segment a scene.

\noindent \textbf{Loss Balancing.}
As in Tab.~\ref{tab:ablation} (3), training with a weak geometry weight degrades the frozen geometric performance while giving lower semantic performance, similarly to the FT semantics only ablations in Tab.~\ref{tab:ablation} (1). This demonstrates that finetuning the pretrained representations on both tasks not only preserves the geometry better, but is also beneficial for segmentation. On the contrary, increasing the geometry weight too much gives marginal gains in MVD, but loses semantic understanding.

\noindent\textbf{Model Components.} We ablate the design choices comparing to the closest method PanSt3R~\cite{zust2025panst3r}. First, as we train %
in-domain, we replace the frozen text embeddings with a (learnable) dense layer. Tab.~\ref{tab:ablation_components} shows that learning the class embeddings yields diminishing returns in segmentation scores. In our method, the performance bottleneck is more the instance detection and refining boundaries rather than classifying a given mask. To investigate the impact of the image encoder, we also experiment with adding %
DINOv2~\cite{oquab2023dinov2} to our method. Simply inputting DINOv2 %
features to the mask decoder makes our approach %
heavier %
while degrading
the segmentation results. Finally, we apply QUBO~\cite{zust2025panst3r} on our predictions with the default parameters of PanSt3R~\cite{zust2025panst3r}, which gives small regression in segmentation scores, therefore adding processing time without benefit.

\input{tables/ablation_model_components}
\input{tables/gradient_weighting}

\noindent\textbf{Gradient Scaling.} %
We take the Pi3~\cite{wang2025pi} model that we train and evaluate in ScanNet++, following the same frame selection protocol as defined at the beginning of this section for ScanNet. We train the model and experiment with %
different scaling factors ranging from $0.1$ (downscaling segmentation gradients) to $2.0$ (multiplying the segmentation gradients) that we apply before the gradient propagation in the multi-view decoder. Results on segmentation, per-view depth and MVD in Tab.~\ref{tab:gradient_ablation} indicate that reducing the semantic gradients in the geometry decoder yields slightly lower segmentation, but improves geometry the most. Even though not scaling (factor $1.0$) improves both geometry and semantics contrary to $2.0$ and $0.5$, we retain for our final model the $0.1$ factor that produces good segmentation (-1 point in mIoU) and a bigger improvement in geometry (+3\% in per view and multiview inliers).

\subsection{Qualitative Observations}
\label{sec:exp:qualitative}

\begin{figure}[t] %
    \centering
    \includegraphics[width=1.0\textwidth]{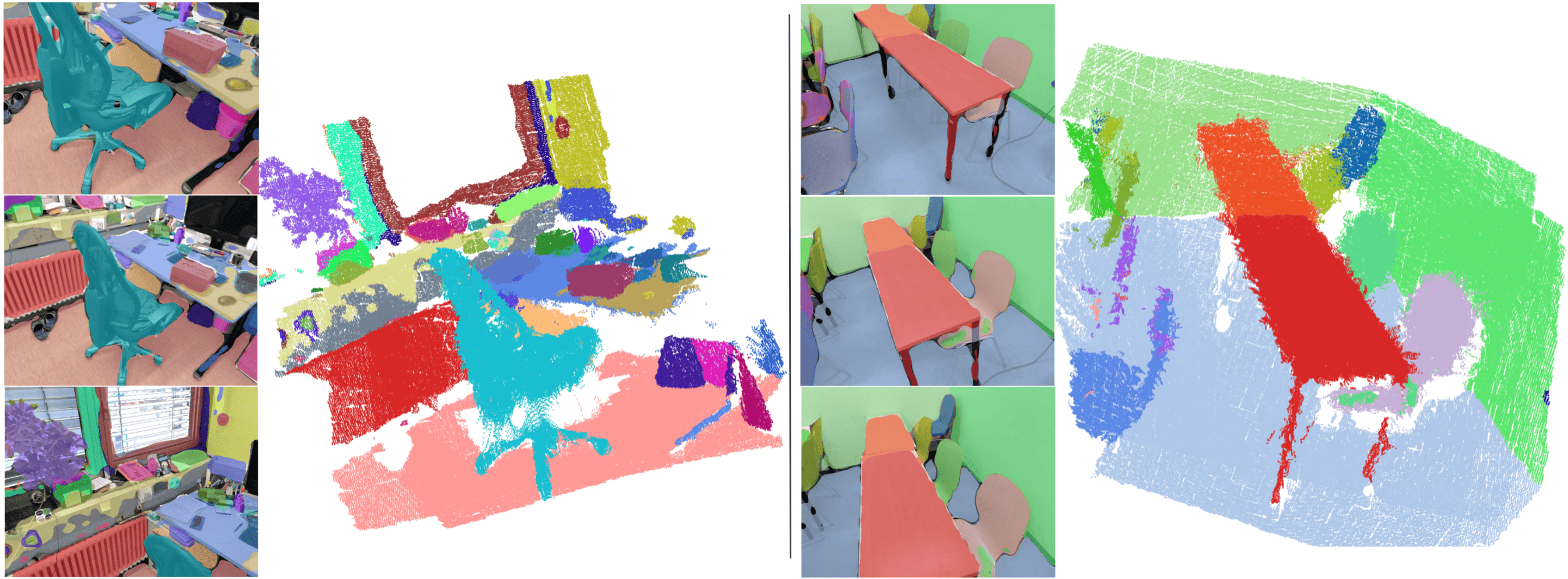}
    \vspace{-.3cm}
    \caption{\textbf{%
    3D reconstruction and instance segmentation  on ScanNet++:} Our model detects 3D-consistent instances across a given sequence (the same query highlights the same object in 3D in all the views). Instances are overlaid on the RGB images, where one color corresponds to one mask (as there can be many instances, some colors may be close). Our model is able to detect small objects, and to maintain good reconstruction capabilities in a shared reference frame in a single feedforward pass.}
    \vspace{-.3cm}
    \label{fig:qualitative_scannetpp}
\end{figure}

\textbf{3D Semantic Instance Segmentation.} Fig.~\ref{fig:qualitative_scannetpp} demonstrates the capabilities of our Pano3D (MUSt3R\cite{cabon2025must3r} backbone) on ScanNet++~\cite{yeshwanth2023scannet++}. In challenging scenes with a lot of small objects (Fig.~\ref{fig:qualitative_scannetpp} left), the queries successfully detect the same object throughout the input sequence. In environments with similar objects (like the chairs or the two nearby tables on the right of Fig.~\ref{fig:qualitative_scannetpp} right), the model correctly %
separates the objects.
	Predicting a pointmap that is pixel-aligned with the masks enables the direct extraction of a segmented 3D point cloud in a single feedforward pass.
We provide additional qualitative results with different backbones in \appendixref{Sec.~C of the supplementary}{Appendix~\ref{sec:supp:qualitative}}.

\begin{figure}[t]
  \centering
  \includegraphics[width=1.0\textwidth]{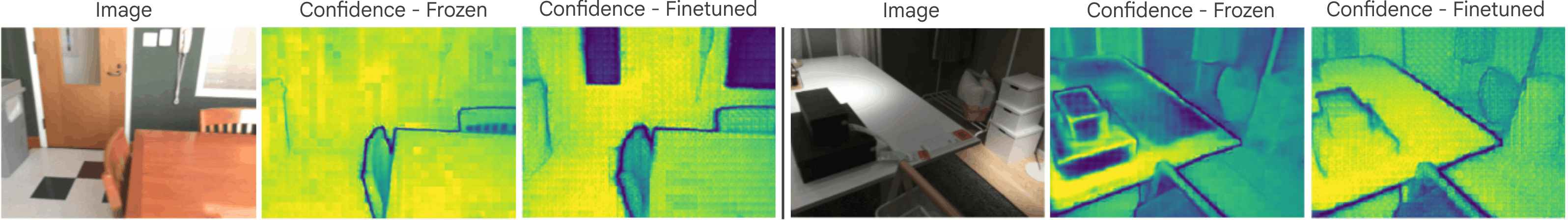}
  \vspace{-.2cm}
  \caption{\textbf{Benefits of joint training:} The middle column shows the confidence maps from the frozen MUSt3R~\cite{cabon2025must3r} model. The model doesn't detect any anomaly on the wall with windows (on the left example) and is unconfident on reflective areas (like the big table on the right). Our model, finetuned with geometry and segmentation supervision, shows a different behavior that demonstrates how joint training modifies the geometric understanding. The finetuned model does not amplify the confidence of the frozen model, but demonstrates object understanding by highlighting glass windows as ambiguous areas separately from the door or wall (left example), and reinforcing the confidence on the center of objects despite shadows (on the boxes on the right example) or reflections (on the table).
  }
  \vspace{-.3cm}
  \label{fig:joint_finetuning}
\end{figure}

\noindent\textbf{Benefits of Joint Training.} Optimizing the geometry decoder modifies the feature space by injecting `objectness' priors. We %
look at the modifications of this joint training on geometry
by visualising the confidence associated to the 3D predictions of the geometry backbone.
As shown in Fig.~\ref{fig:joint_finetuning}, the frozen MUSt3R backbone hallucinates confidence on ambiguous geometries (e.g., treating a glass door as a wall on the left example) and is unconfident on reflective, textureless areas (right example). On the contrary, the finetuned model detects ambiguous areas like windows and is legitimately confident on objects like tables and boxes. More largely, we observe that the uncertainty in the finetuned model is mostly at the edges of objects, and on ambiguous areas (windows, mirrors).

\begin{figure}[t] %
    \centering
    \includegraphics[width=1.0\textwidth]{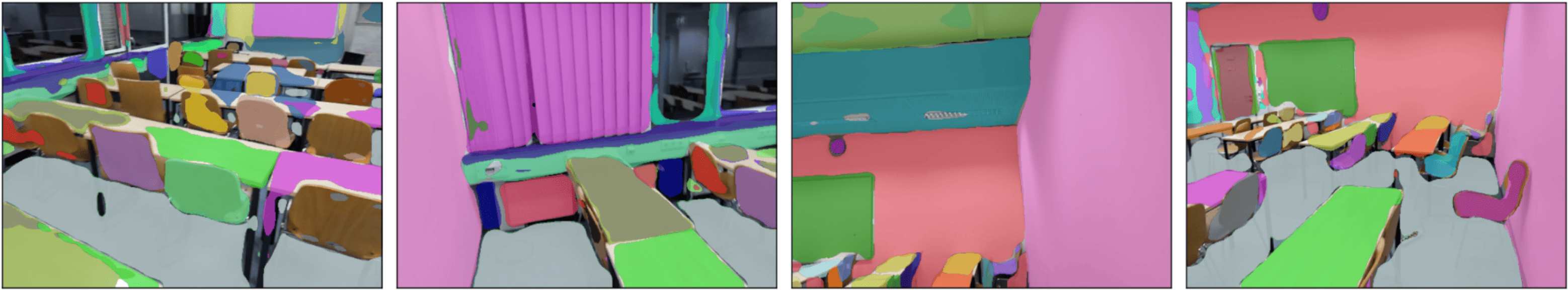}
    \vspace{-.4cm}
    \caption{\textbf{Cluttered environments:} In scenes where there are a lot of objects, %
    our model struggles to separate instances and groups different objects into one mask. Here, all the frames are part of the same sequence. We overlay instance predictions in different colors (due to the high number of instances, some might share similar colors). Some masks in the first image (left) group two different chairs or two different tables.}
    \vspace{-.3cm}
    \label{fig:failure_clutter}
\end{figure}

\noindent \textbf{Cluttered Environments.} Our method also inherits the drawbacks of query-based detection. In environments with a lot of objects as in Fig.~\ref{fig:failure_clutter}, our model struggles to detect all the instances and detections `bleed' on multiple objects. For instance, one query might detect two chairs that are close.

%% file: tables/ablations.tex
\begin{table} [t] %
\centering
\caption{\textbf{Ablation studies on ScanNet validation.} We investigate training regimes, architectural components, external priors, and backbone generality. (FT: Finetuned, G: Geometry, S: Semantics). Joint optimization (\textbf{G+S})
maintains geometric fidelity while improving instance localization.}
\vspace{-.3cm}
\label{tab:ablation}
\resizebox{\textwidth}{!}{%
\setlength{\tabcolsep}{4pt} %
\begin{tabular}{l cc cc cc}
\toprule
 & \multicolumn{2}{c}{\textbf{3D Mesh}} & \multicolumn{2}{c}{\textbf{2D Metrics}} & \multicolumn{2}{c}{\textbf{Multi View Depth}} \\
\cmidrule(lr){2-3} \cmidrule(lr){4-5} \cmidrule(lr){6-7}
\textbf{Variant} & \textbf{mIoU3D} $\uparrow$ & \textbf{AP3D} $\uparrow$ & \textbf{mIoU2D} $\uparrow$ & \textbf{mAcc2D} $\uparrow$ & \textbf{AbsRel} $\downarrow$ & \textbf{$\tau < 1.03$} $\uparrow$ \\
\midrule
\textit{(1) Training Regime (MUSt3R)} \\
Frozen Geo + Sem Head & 61.9 & 10.8 & 72.1 & 83.1 & 5.2 & 44.8\% \\
FT Geometry Only (G)  & n/a & n/a & n/a & n/a & 3.4 & 58.1\% \\
FT Semantics Only (S) & n/a & n/a & 72.0 & 82.8 & 156.3 & 2.0\% \\
End to end (training encoder) & 63.4 & \textbf{16.6} & 72.8 & 82.9 & \textbf{3.3} & \textbf{60.7}\% \\
\rowcolor[gray]{0.9} \textbf{Pano3D (MUSt3R)} (G + S) & \textbf{65.3} & 15.7 & \textbf{73.5} & \textbf{83.7} & 3.5 & 57.2\% \\
\midrule
\textit{(2) Feature Bridge} \\
Encoder Context ($E_i$) only & 60.4 & 7.8 & 70.5 & 82.0 & \textbf{3.4} & \textbf{58.1\%} \\
Final Geo features ($F_i$) only & 63.8 & 13.9 & 73.0 & \textbf{83.9} & 3.7 & 53.2\% \\
\rowcolor[gray]{0.9} \textbf{Pano3D (MUSt3R)} ($F_i$ + $E_i$) & \textbf{65.3} & \textbf{15.7} & \textbf{73.5} & 83.7 & 3.5 & 57.2\% \\
\midrule
\textit{(3) Loss coefficient} \\
$\lambda_{geo}$=$\lambda_{sem}$ & 61.8 & 12.1 & 72.9 & 82.1 & 5.1 & 39.6\% \\
$\lambda_{geo}$=$20 \times \lambda_{sem}$ & 63.3 & \textbf{16.2} & 71.7 & 81.8 & \textbf{3.4} & \textbf{57.0\%} \\
\rowcolor[gray]{0.9} \textbf{Pano3D (MUSt3R)} ($\lambda_{geo}$=$10 \times \lambda_{sem}$) & \textbf{65.3} & 15.7 & \textbf{73.5} & \textbf{83.3} & 3.5 & 56.7\% \\
\bottomrule
\end{tabular}
}
\end{table}

%% file: tables/ablation_model_components.tex
\begin{table} [t] %
    \centering
    \caption{\textbf{Ablation studies on model components (ScanNet).}}
    \vspace{-.3cm}
    \label{tab:ablation_components}
    \begin{tabular}{l cc}
        \toprule
        \textbf{Variant} & \textbf{mIoU2D} $\uparrow$ & \textbf{mAcc2D} $\uparrow$ \\
        \midrule
        \rowcolor[gray]{0.9} Pano3D & 73.5 & 83.7 \\
        \quad Pano3D + DinoV2 & 71.7 & 82.5 \\
        \quad Pano3D + Dense Head & \textbf{73.9} & \textbf{84.2} \\
        \quad Pano3D + Dense Head and QUBO~\cite{zust2025panst3r} & 72.6 & 83.6 \\
        \bottomrule
    \end{tabular}
    \vspace{-.3cm}
\end{table}

%% file: tables/gradient_weighting.tex
\begin{table} [t] %
    \centering
    \small
    \setlength{\tabcolsep}{5pt} %
    \caption{\textbf{Impact of the gradient scaling.}} %
    \vspace{-.3cm}
    \label{tab:gradient_ablation}
    \begin{tabular}{l cccccc cc}
        \toprule
        & \multicolumn{2}{c}{Segmentation} & \multicolumn{2}{c}{Per-Frame Depth} & \multicolumn{2}{c}{MVD} \\
        \cmidrule(lr){2-3} \cmidrule(lr){4-5} \cmidrule(lr){6-7}
        Gradient Factor & mAcc & mIoU & AbsRel & $\tau < 1.03$ & AbsRel & $\tau < 1.03$ \\
        \midrule
        Frozen Pi3~\cite{wang2025pi} & n/a & n/a & 2.58 & 81.25 & 3.01 & 73.11 \\
        1.0 (no scaling)                  & 61.59 & 47.41 & 2.44 & 82.99 & 2.95 & 73.20 \\
        2.0                   & 60.50 & 44.86 & 2.49 & 82.32 & 3.18 & 68.93 \\
        0.5                   & \textbf{61.79} & \textbf{47.53} & 2.42 & 83.99 & 3.29 & 66.28 \\
        0.1                   & 60.50 & 44.91 & \textbf{2.32} & \textbf{84.56} & \textbf{2.77} & \textbf{76.26} \\
        \bottomrule
    \end{tabular}
    \vspace{-.3cm}
\end{table}

%% file: sections/06_conclusion.tex
\section{Conclusion}
\label{sec:conclusion}

We presented Pano3D, a streamlined integrated framework for joint 3D reconstruction and panoptic segmentation from unposed image collections. 
By appending a set-based panoptic head to the final representations of feedforward reconstruction models and jointly finetuning the entire system, Pano3D manages to combine information at different stages and achieves state-of-the-art results. This work opens a path towards treating 3D reconstruction and scene understanding from unposed views as one task. While our current method relies on finetuning an existing geometry-first backbone, our findings suggest that potential future foundation models should be trained jointly on structure and semantics from inception. We envision fully unified representations where scene-level geometric information interacts with object centric representations.

%% file: sections/07_acknowledgements.tex
{ \small
	\noindent\textbf{Acknowledgements.}
	We acknowledge Walid Bousselham, Mohammad Fahes and Théodore Fougereux for the feedback and support throughout the project. We also thank Anurag Arnab and Guillaume Le Moing for the insightful discussions.
}

%% file: supp_mat.tex
\beginsupplement

In this supplementary document, we provide additional technical details, expanded experimental results, and visualizations. Specifically:
\begin{itemize}
    \item Section~\ref{sec:supp:implem}: Implementation details and training setting.
    \item Section~\ref{sec:supp:ablations}: Additional ablation experiments.
    \item Section~\ref{sec:supp:qualitative}: Qualitative results.
    \item Section~\ref{sec:supp:experiments}: Additional experiments
\end{itemize}

\input{supp_mat/S00_implem_details}
\input{supp_mat/S01_ablations}
\input{supp_mat/S02_qualitative}   %
\input{supp_mat/S03_experiments}   %

%% file: supp_mat/S00_implem_details.tex
\section{Implementation Details}
\label{sec:supp:implem}

\noindent\textbf{Architecture.}
We initialize the weights from MUSt3R~\cite{cabon2025must3r}. It consists of a finetuned CroCov2~\cite{weinzaepfel2023croco} ViT-L encoder and a ViT-B cross-view decoder. We extract the encoder and decoder features, and process them to create the frame features $\mathcal{G}_i$ that are used as input to the mask decoder, which is initialized from \cite{cheng2022masked}. However, the object queries don't attend to multiscale features, but only the patch-level $\mathcal{G}_i$ to save computation. Weights of the mask decoder are initialized from pretrained Mask2Former (trained on COCO~\cite{lin2014microsoft_coco} with a ResNet50~\cite{he2016deep_resnet} backbone). 

The frame features $\mathcal{G}_i$ are first passed in a standard MLP with GeLU~\cite{hendrycks2016gaussian_gelu} activations. These features are then used as keys and values in cross-attention with object queries in the mask decoder, but they are also used to produce high-level mask features $\mathcal{M}_i$. To that end, they successively go through MLPs and Pixel Shuffle operations to be upsampled from stride 16 (patch level) to stride 4. For instance, for a 384x512 input, our mask predictions are at a resolution 96x128. The 3D head from MUSt3R is kept identical as in the original model~\cite{cabon2025must3r}: it is a linear head that takes as input the final decoder features and predicts dense information at input resolution. For each pixel, 7 values are predicted: confidence (1 value), pointmap in local coordinate frame (3 values), and pointmap in global coordinate frame (3 values). Finally, to lift the predictions to 3D, we bilinearly resize the predicted mask logits to the input resolution to get a direct pixel to pixel correspondence between the geometric predictions and segmentation predictions. Same goes for Pi3~\cite{wang2025pi}: we reuse the model heads described from the original article. %

\vspace{.2cm}
\noindent\textbf{Geometric loss.}
We introduced two backbones for Pano3D, namely MUSt3R~\cite{cabon2025must3r} and Pi3~\cite{wang2025pi}. They share a similar task (predict global geometry), but the objectives differ. MUSt3R uses the first frame as the reference frame, and for each frame $f$ regress a global pointmap (in the reference frame's coordinate system) as well as local pointmap (in $f$ coordinate system). It is put to log space to be more stable on long scenes, and weighted by confidence as in DUSt3R~\cite{wang2024dust3r}. Pi3~\cite{wang2025pi} gets rid of the reference frame by predicting scale-invariant local pointmaps pairwise and affine-invariant camera poses (composed of an angle loss and a Huber loss). On top of this, the authors added a normal loss (for surface smoothness) and a confidence loss (Binary Cross Entropy). Specifics are explained in the dedicated method sections of these approaches\cite{cabon2025must3r, wang2025pi}.

\begin{figure}[t] %
    \centering
    \includegraphics[width=1.0\textwidth]{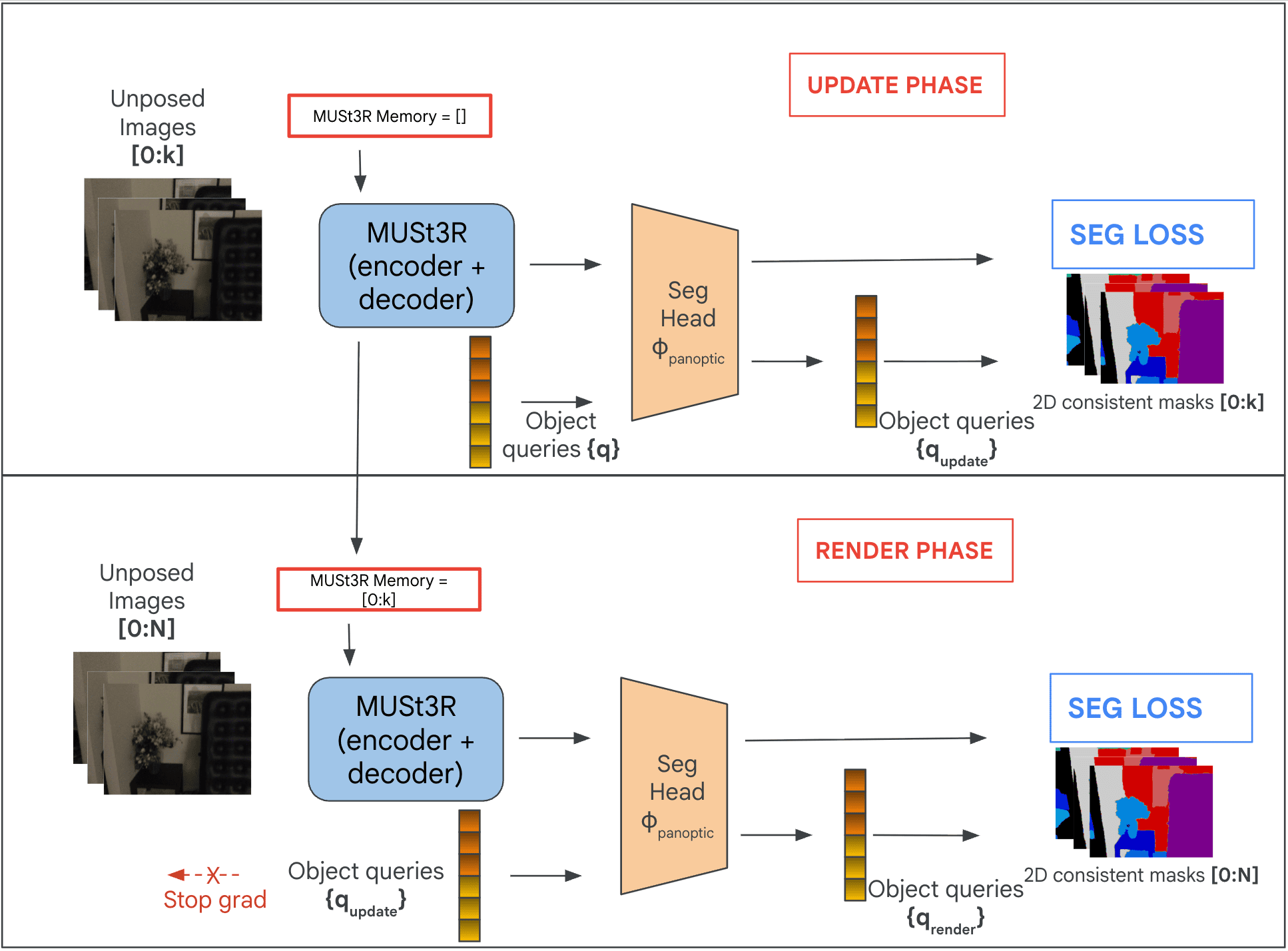}
    \vspace{-.4cm}
    \caption{\textbf{Training consistent queries with MUSt3R~\cite{cabon2025must3r} memory:} We reuse the queries between the update (top block) and render (bottom block) with a stop gradient mechanism. We then supervise the predictions across both phases with the same ground truth batch (we mask the non selected frames in the update phase). With this mechanism, we ensure that the object queries identities remain identical between both phases.
    }
    \label{fig:supp:query_tracking}
\end{figure}

\vspace{.2cm}
\noindent\textbf{Training Setting for the Online Model.} As explained in Sec.~3.4, we supervise our model in two phases. This training scheme is inspired from MUSt3R~\cite{cabon2025must3r} and we adapt it to enforce that object queries in the update phase (where we fill the memory) and the render phase (not adding to memory) keep the same identity. We provide a detailed figure of the mechanism in Fig.~\ref{fig:supp:query_tracking}. During training, we randomly sample $k$ frames from the batch for the update phase, and input all the frames in the batch for the render phase (ensuring each frame has minimal overlap with an update frame). The queries in the render phase start from the final state of the update phase, but we stop the gradients between the two phases. Enabling the gradient flow between the phases would create an information leak where new frames in the render phase (those not selected in the $k$ frames) could influence the update phase. We supervise mask predictions from both phases altogether by tiling the batch along batch dimension. That way, the mask predictions for object query $i$ are forced to be consistent across phases.

\vspace{.2cm}
\noindent\textbf{Training Schedule.}
All models are trained on 8 TPU v6 accelerators with a batch size of 1 per device using the AdamW optimizer~\cite{kingma2017adammethodstochasticoptimization} with gradient clipping at norm 1.0. We use a compound learning rate schedule with linear warmup (3000 steps) followed by multiplicative decay at intervals of 30--95\% of total training. We apply different learning rates per parameter group: the bridge components (adapters, fusion, upscaler) use the base rate $1\mathrm{e}{-4}$, while segmentation heads and geometry decoder use $1\mathrm{e}{-4}$. The segmentation parameters also have a weight decay of 0.05. We train for 30k steps on the ScanNet~\cite{dai2017scannet} dataset, and 50k steps for ScanNet200~\cite{rozenberszki2022language_scannet200} and ScanNet++~\cite{yeshwanth2023scannet++} that require more training to converge due to the higher number of classes. %

\vspace{.2cm}
\noindent\textbf{Inference.}
We benchmark inference time of our approach with different backbones MUSt3R~\cite{cabon2025must3r} and Pi3~\cite{wang2025pi} on a single H100 GPU for a sequence of 100 frames at resolution 512. The model sizes and average time over 5 runs is reported in is in Tab.~\ref{tab:supp:model_specs}. %

\vspace{.2cm}
\noindent\textbf{Dataset.}
To create the training splits for our three datasets~\cite{dai2017scannet, rozenberszki2022language_scannet200, yeshwanth2023scannet++}, we first subsample every 20 frames and then randomly select a starting view $v_{\text{start}}$. Then, we randomly sample a step in $s \in [1, 4]$ and sample a sequence of 15 frames starting from $v_{\text{start}}$. The final sequence is defined as:
\[
\{v_{\text{start}}, v_{\text{start}+s}, v_{\text{start}+2s}, \dots, v_{\text{start}+15s}\}
\]
This data is preprocessed and used by the dataloader. During training, if a view in the render phase has no overlap with one of the update views, we mask its predictions so it does not penalize the loss.

%% file: supp_mat/S01_ablations.tex
\section{Additional Ablations}
\label{sec:supp:ablations}

\noindent\textbf{Number of Queries.} We study for each dataset how the number of object queries impacts the segmentation performance. To do so, we take the frozen feedforward geometry backbone and train the semantic head (bridge+mask decoder) with either 100, 200 or 400 queries and evaluate it on the UNITE benchmark for 20k steps on the
ScanNet++~\cite{yeshwanth2023scannet++} (100 classes) dataset.

\input{tables/supp_ablations_numqueries}

Results in Tab.~\ref{tab:supp:numqueries} suggest that training with more queries increases the performance on datasets with a high number of classes.
Having 200 queries yields better segmentation on ScanNet++~\cite{yeshwanth2023scannet++} dataset, since there are more objects per scene in average. We thus try to increase the number queries to 400 on ScanNet++ and observe diminishing returns, pointing out to the saturation effect also evidenced in~\cite{cheng2022masked}. The full ablation on the fully finetuned Pano3D on all datasets is left to future work, but the ScanNet++ results give similar conclusions as Tab.~\ref{tab:supp:numqueries}.

\vspace{.2cm}
\noindent\textbf{Segmenting Directly in 3D.} A natural approach for 3D semantic segmentation from unposed views would be to have a 2-stage pipeline. First, we predict the 3D pointcloud using a feedforward reconstruction model, and then give it as input to a segmentation model that takes posed RGB-D as input. We reimplement ODIN~\cite{Jain2024ODINAS} that is, at the time of writing, the state-of-the-art on ScanNet 3D semantic segmentation benchmark. ODIN alternates 2D blocks from a classical backbone (ResNet50~\cite{he2016deep_resnet} or Swin~\cite{liu2021swin}) and 3D blocks that consist in projecting the frames to 3D and perform cross attention with k-nearest neighbors in 3D space (typically, $k=8$). It then uses a similar mask decoder to Pano3D where object queries predict mask over the full video sequence, but use a learned 3D positional encoding added to the input features. We reimplement the ResNet50 variant and compare the 3D mIoU in Tab.~\ref{tab:supp:odin}. One detail though is that ODIN supervises directly in 3D, and thus interpolates the RGB-D pointcloud on the official mesh (which is annotated). In this experiment, we predict the pointcloud in the reference frame on our input sequences ScanNet~\cite{dai2017scannet} sequences with pretrained MUSt3R~\cite{cabon2025must3r} and the ground truth reference camera pose to align it to the world reference. We thus have pseudo-ground truth posed RGB-D sequences. We use this as input to ODIN~\cite{Jain2024ODINAS}, train the model and evaluate 3D mIoU in Tab.~\ref{tab:supp:odin}. 

\input{tables/supp_ablations_odin}

Even without using the ground truth depth and camera information to lift the predictions, our Pano3D gets $56.1$ mIoU, which is still below ODIN~\cite{Jain2024ODINAS} by more than 10 points. In contrast, when addressing the 3D segmentation problem as first reconstruct, then segment, the ODIN~\cite{Jain2024ODINAS} model fails in establishing correspondence with the noisy pointcloud.

%% file: tables/supp_ablations_numqueries.tex
\begin{table}[t] %
\centering
\caption{\textbf{Ablation on the Number of Queries.} We train Pano3D with 100, 200, or 400 object queries and measure the mIoU3D and AP3D on the UNITE~\cite{koch2025unified} benchmark. We train under the same conditions: the geometry backbone is frozen and only the segmentation head is trained.}
\vspace{-.3cm}
\label{tab:supp:numqueries}
\setlength{\tabcolsep}{8pt}
\begin{tabular}{l cc}
\toprule
& \multicolumn{2}{c}{ScanNet++~\cite{yeshwanth2023scannet++}} \\
\cmidrule(lr){2-3}
Variant & mIoU $\uparrow$ & mAcc $\uparrow$ \\
\midrule
Pano3D - 100 queries & 25.3 & 36.9 \\
Pano3D - 200 queries & \textbf{26.8} & 40.1 \\
Pano3D - 400 queries & 26.7 & \textbf{41.7} \\
\bottomrule
\end{tabular}
\end{table}

%% file: tables/supp_ablations_odin.tex
\begin{table} [t] %
\centering
\caption{\textbf{Segmenting Directly in 3D (ScanNet).} We train ODIN~\cite{Jain2024ODINAS} that segments in 3D a posed RGB-D sequence. Then, we term `ODIN+MUSt3R' the variant where, instead of ground-truth depth and camera poses, we use the MUSt3R~\cite{cabon2025must3r} predictions (keeping the model frozen). As the predictions are noisier, the model doesn't manage to establish reliable relationships in 3D, resulting in highly degraded results. $^\dagger$ denotes our reproduction of the model with ResNet backbone using a simple Feature Pyramid Network (FPN)~\cite{lin2017feature_fpn} instead of deformable attention pixel decoder~\cite{cheng2022masked}.}
\vspace{-.3cm}
\label{tab:supp:odin}
\setlength{\tabcolsep}{20pt} %
\begin{tabular}{l c}
\toprule
Variant & mIoU3D $\uparrow$ \\
\midrule
ODIN$^\dagger$ & \textbf{69.3} \\
ODIN$^\dagger$ + MUSt3R & 39.2 \\
Pano3D (Ours) & 56.1 \\
\bottomrule
\end{tabular}
\end{table}

%% file: supp_mat/S02_qualitative.tex
\section{Additional Qualitative Results}
\label{sec:supp:qualitative}

\begin{figure}[t]
  \centering
  \includegraphics[width=\linewidth]{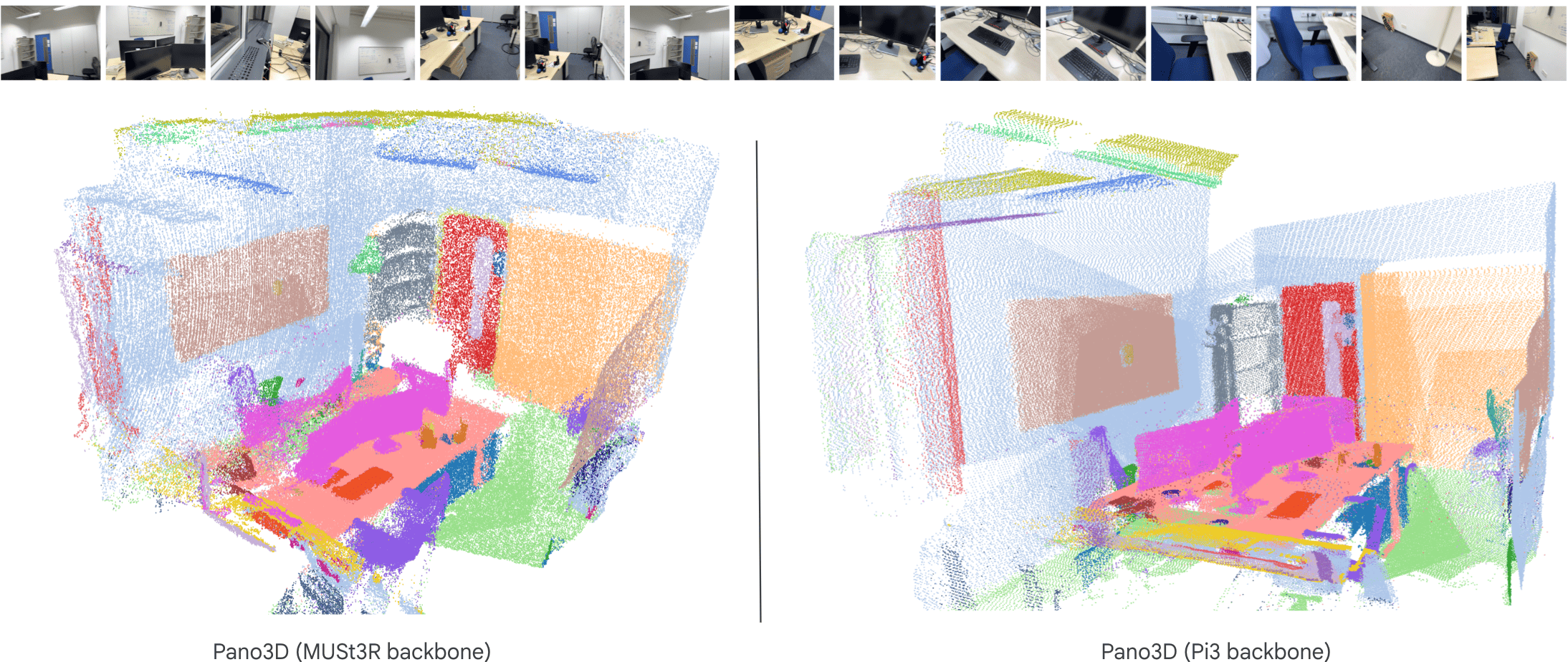} \\
\vspace{.3cm}
  \includegraphics[width=\linewidth]{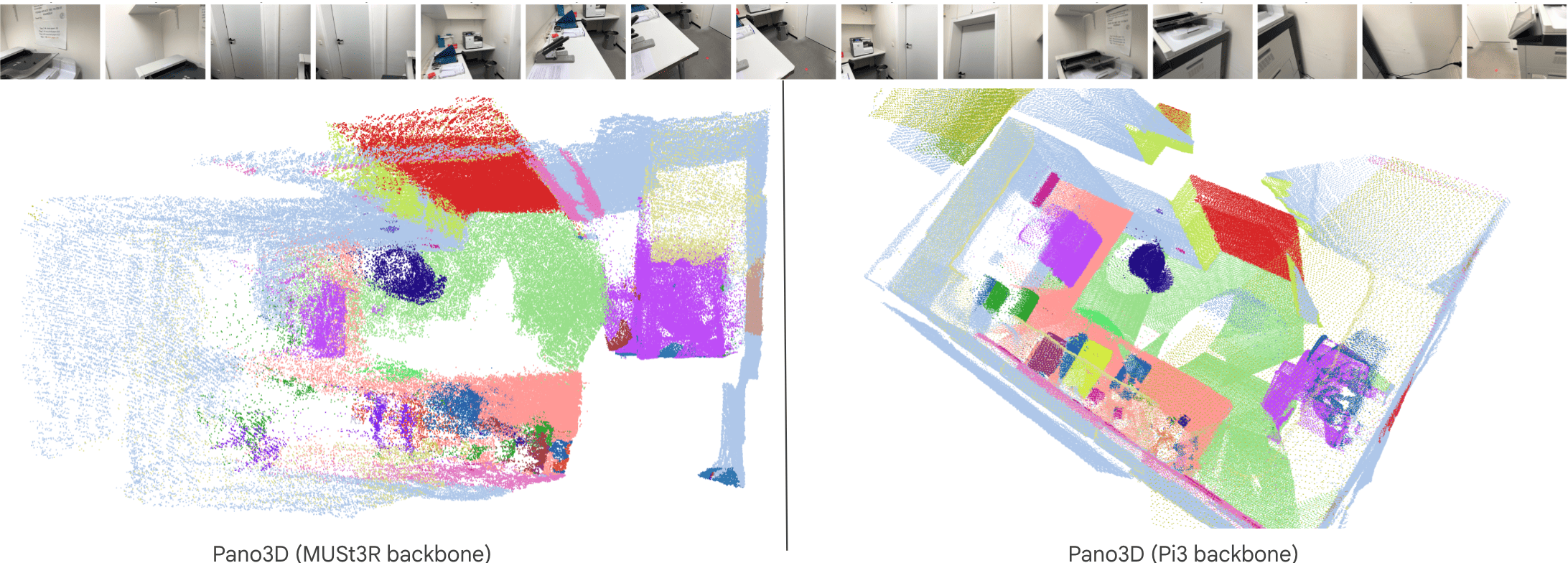}
  \caption{\textbf{Comparison between MUSt3R~\cite{cabon2025must3r} and Pi3~\cite{wang2025pi} backbones with 15 views as input:} \textbf{(top)}
  	Both models manage to reconstruct the scene and segment the objects in a single feedforward pass.
  	\textbf{(bottom)}
  	Pi3 is globally more accurate than MUSt3R. In challenging scenes like this one, Pi3 achieves to reconstruct fine details in the geometry, which has a beneficial impact on the segmentation (detecting objects on the bottom left). MUSt3R which drifts in this case has a lower reconstruction quality, and consequently does not detect all the objects in the scene.}
  \label{fig:supp:comparison}
\end{figure}

\begin{figure}[t]
  \centering
  \includegraphics[width=\linewidth]{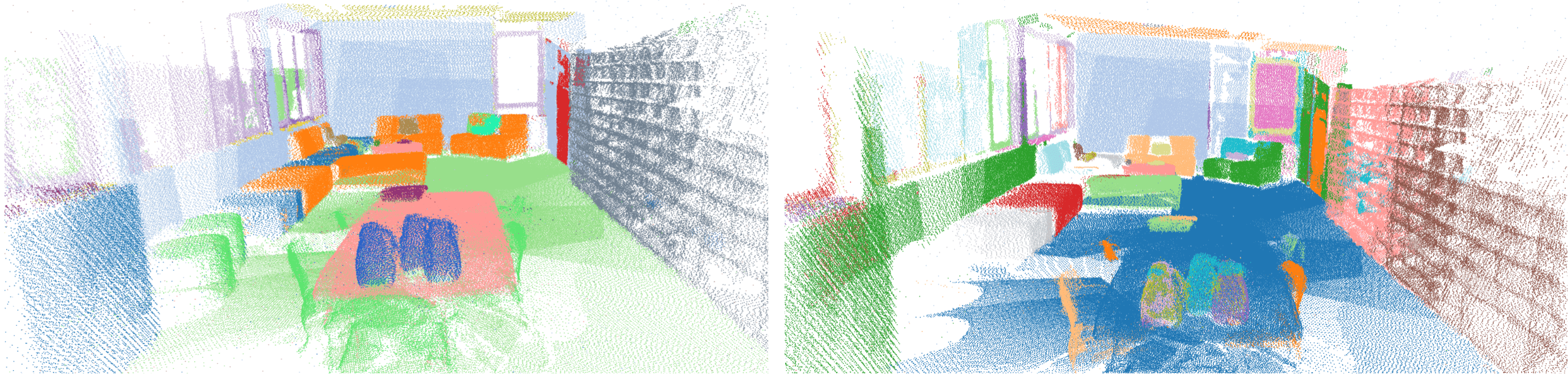}
  \caption{Pano3D (Pi3 backbone) qualitative results in semantic segmentation (left) and instance segmentation (right).}
  \label{fig:supp:combined_sofas_pi3}
\end{figure}

\begin{figure}[t]
  \centering
  \includegraphics[width=\linewidth]{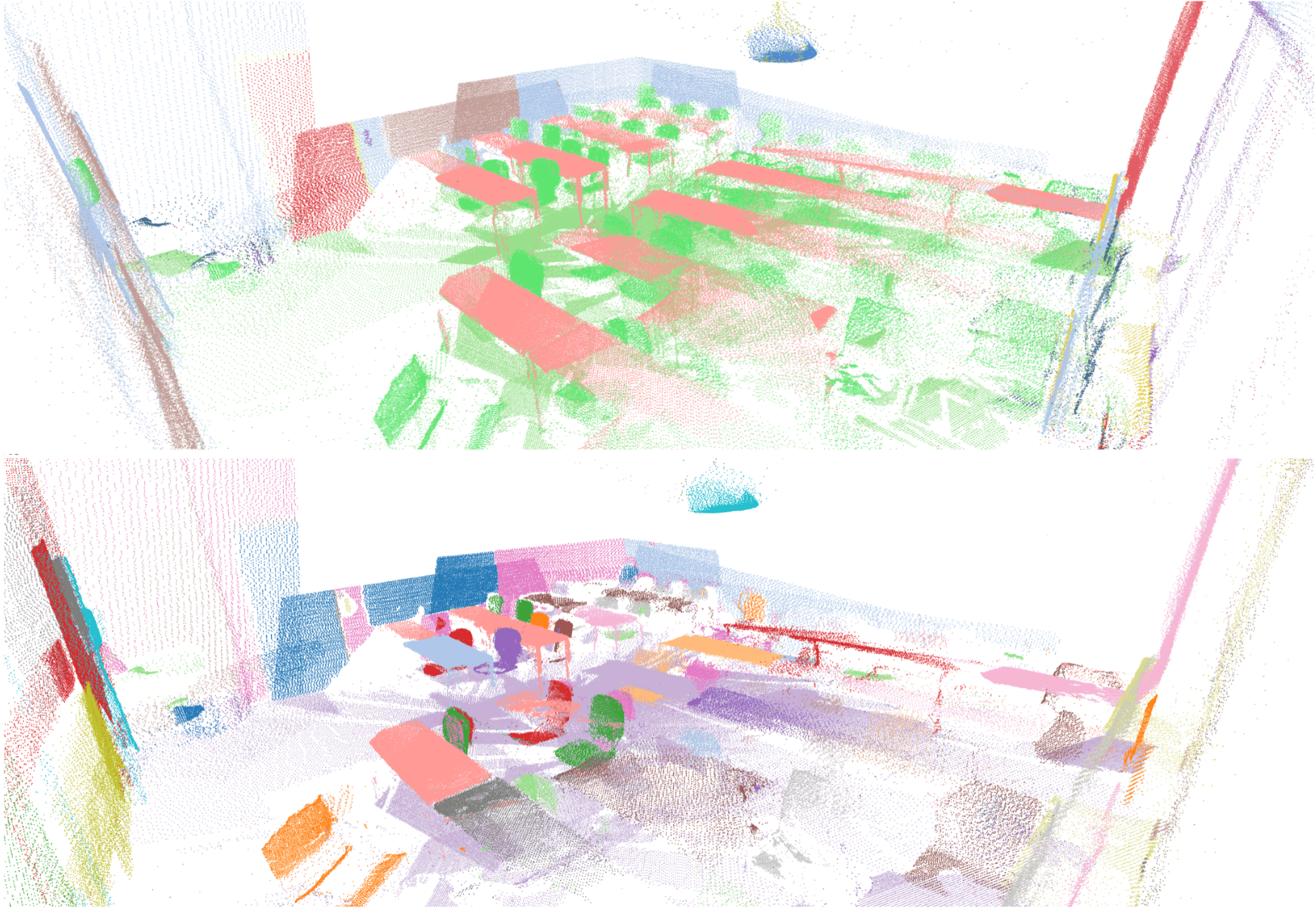}
  \caption{Pano3D (Pi3 backbone) qualitative results in semantic segmentation (top) and instance segmentation (bottom) in a challenging environment with many similar instances.}
  \label{fig:supp:combined_classroom_pi3}
\end{figure}

\noindent\textbf{Comparing Backbones.} Our approach is generalizable to any feedforward reconstruction model as evidenced in Sec.~\ref{sec:supp:ablations}. The model doesn't degrade the underlying quality of the geometric model and accurately captures semantics. Fig.~\ref{fig:supp:comparison}(top)
shows a qualitative comparison between the two backbones. Both accurately reconstruct and segment the scene. However, in more challenging classes, the superiority of the underlying geometric backbone impacts the segmentation quality, as shown in Fig.~\ref{fig:supp:comparison}(bottom).
We provide qualitative results with the  Pi3 backbone in Fig.~\ref{fig:supp:combined_sofas_pi3} and Fig.~\ref{fig:supp:combined_classroom_pi3}.

\vspace{.2cm}
\noindent\textbf{Open-Vocabulary Queries.} To predict the class for a given mask proposal, we use a matrix of frozen text embeddings extracted with CLIP~\cite{radford2021learning} instead of a standard Dense layer. This enables open-vocabulary queries at inference. To do so, we simply replace the matrix with the text embeddings of the classes that we aim to segment. Fig.~\ref{fig:supp:openvoc_queries} shows some results where objects unseen during training are queried. Our model is able to detect areas relevant to the input query in a zero-shot fashion as long as it stays close enough to the training set. Classes or expressions such as `parquet', `rice cooker', `chop vegetables' work well, %
but out-of-distribution classes such as %
`tree' fail, %
since we train %
Pano3D on indoor datasets (where `window' will be detected instead of `tree' and the 3D predictions are unconfident in these ambiguous areas). Training on a vast amount of data that covers more diverse vocabulary is left to future work.

\begin{figure}[p] %
  \centering
  \begin{subfigure}{\linewidth}
    \centering
    \includegraphics[height=1.0\textheight, width=\linewidth, keepaspectratio]{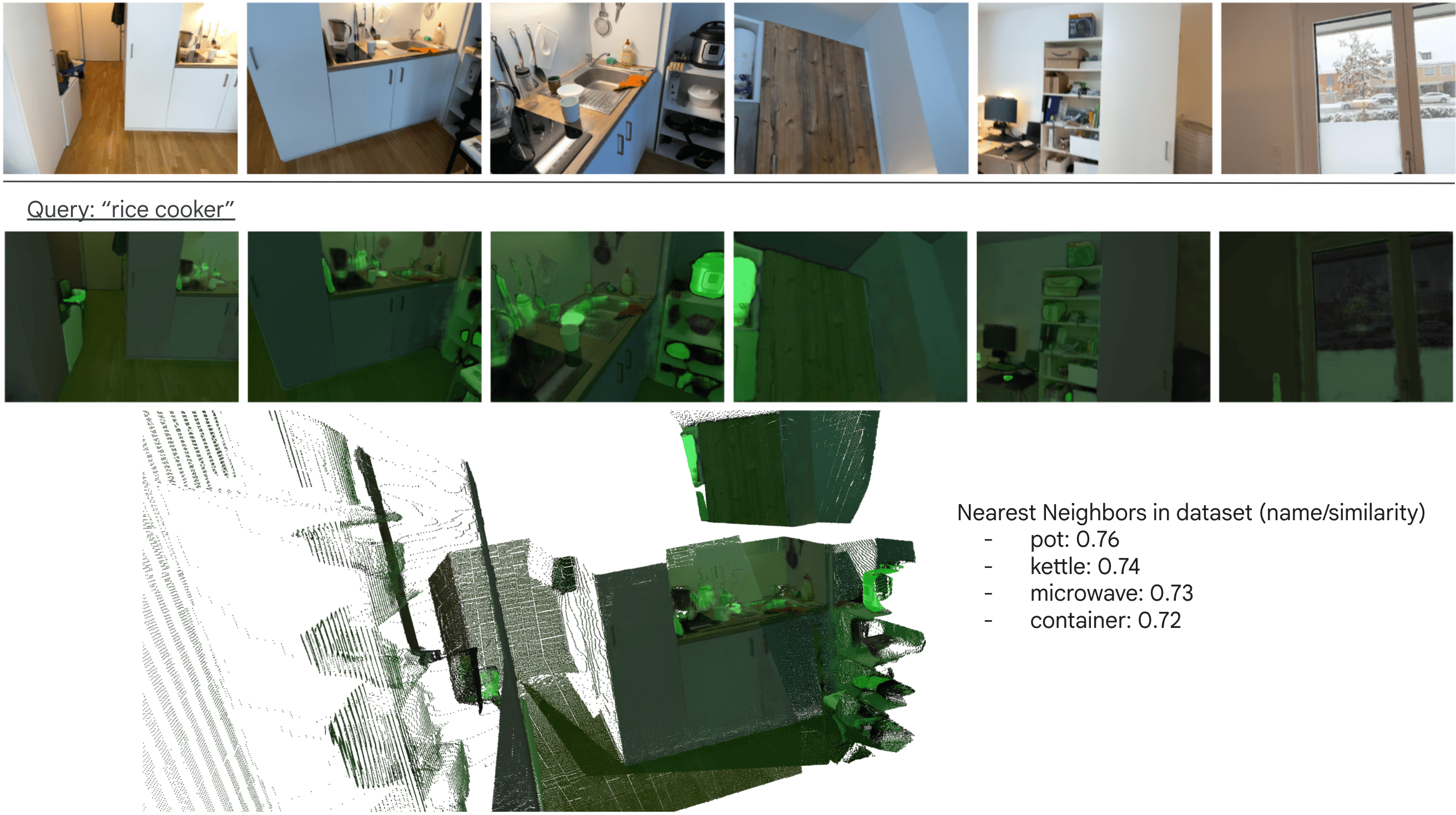}
    \vspace{-5pt}
  \end{subfigure}

  \begin{subfigure}{\linewidth}
    \centering
    \includegraphics[height=1.0\textheight, width=\linewidth, keepaspectratio]{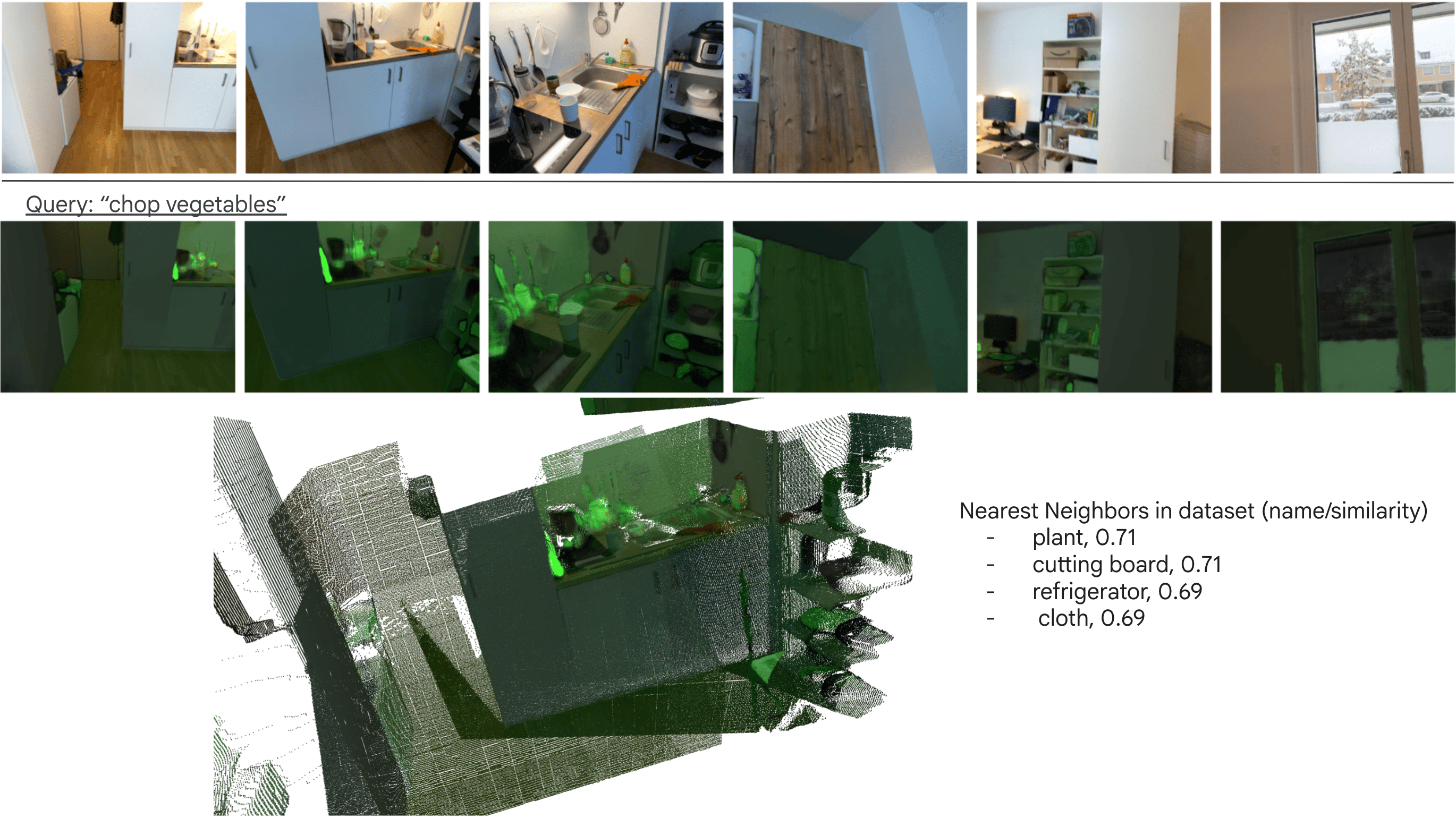}
    \vspace{-5pt}
  \end{subfigure}

  \caption{\textbf{Open-vocabulary queries:} We detect classes not present in the initial training set by replacing text embeddings with new class target embeddings. \textbf{Top to bottom per subfigure:} Input RGB sequence chunk, 2D relevance score (green=relevant), and 3D lifted relevance score.}
  \label{fig:supp:openvoc_queries}
\end{figure}

\begin{figure}[p] %
  \centering
  \begin{subfigure}{\linewidth}
    \centering
    \includegraphics[height=1.0\textheight, width=\linewidth, keepaspectratio]{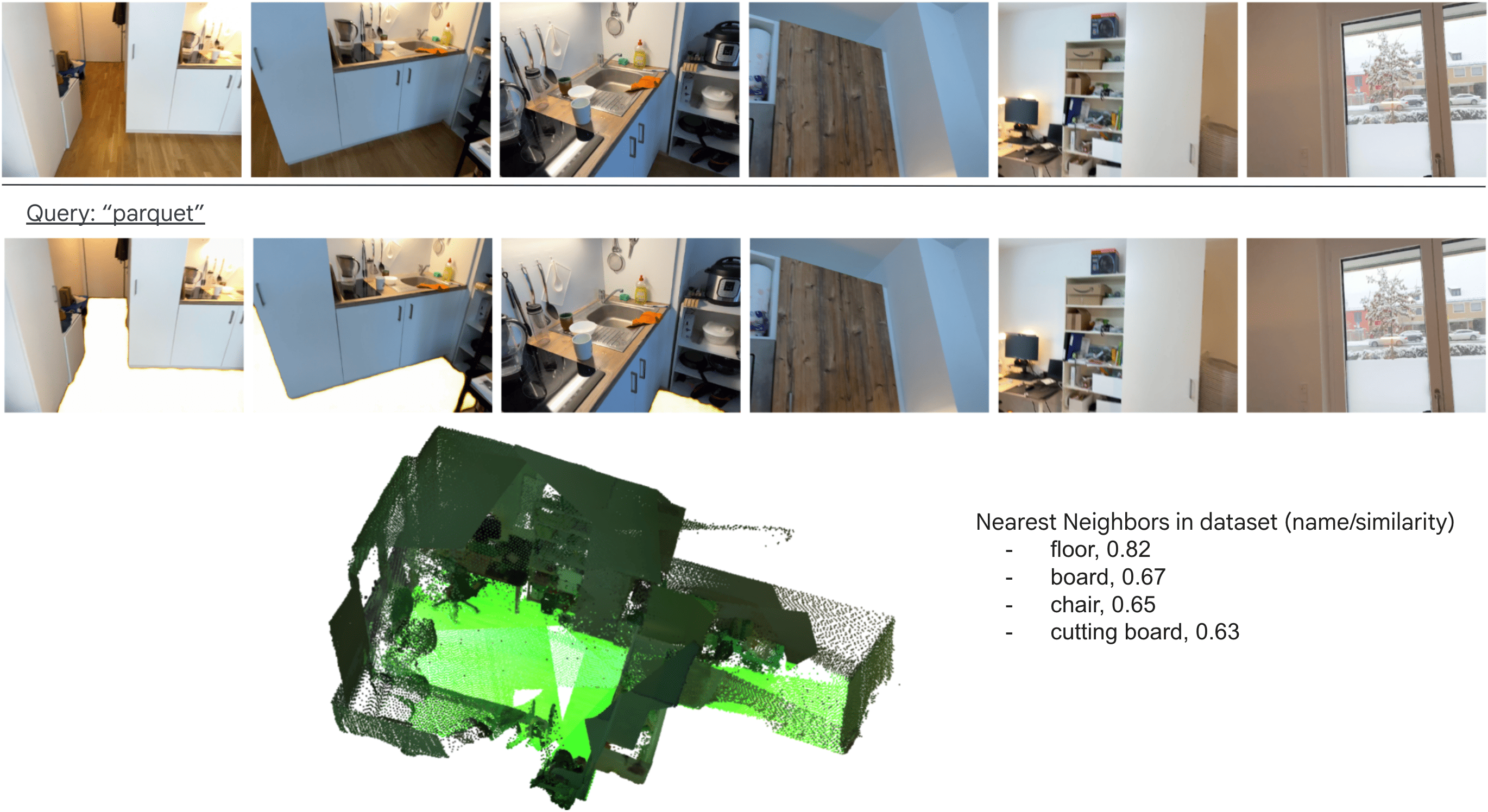}
    \vspace{-5pt} %
  \end{subfigure}

  \begin{subfigure}{\linewidth}
    \centering
    \includegraphics[height=1.0\textheight, width=\linewidth, keepaspectratio]{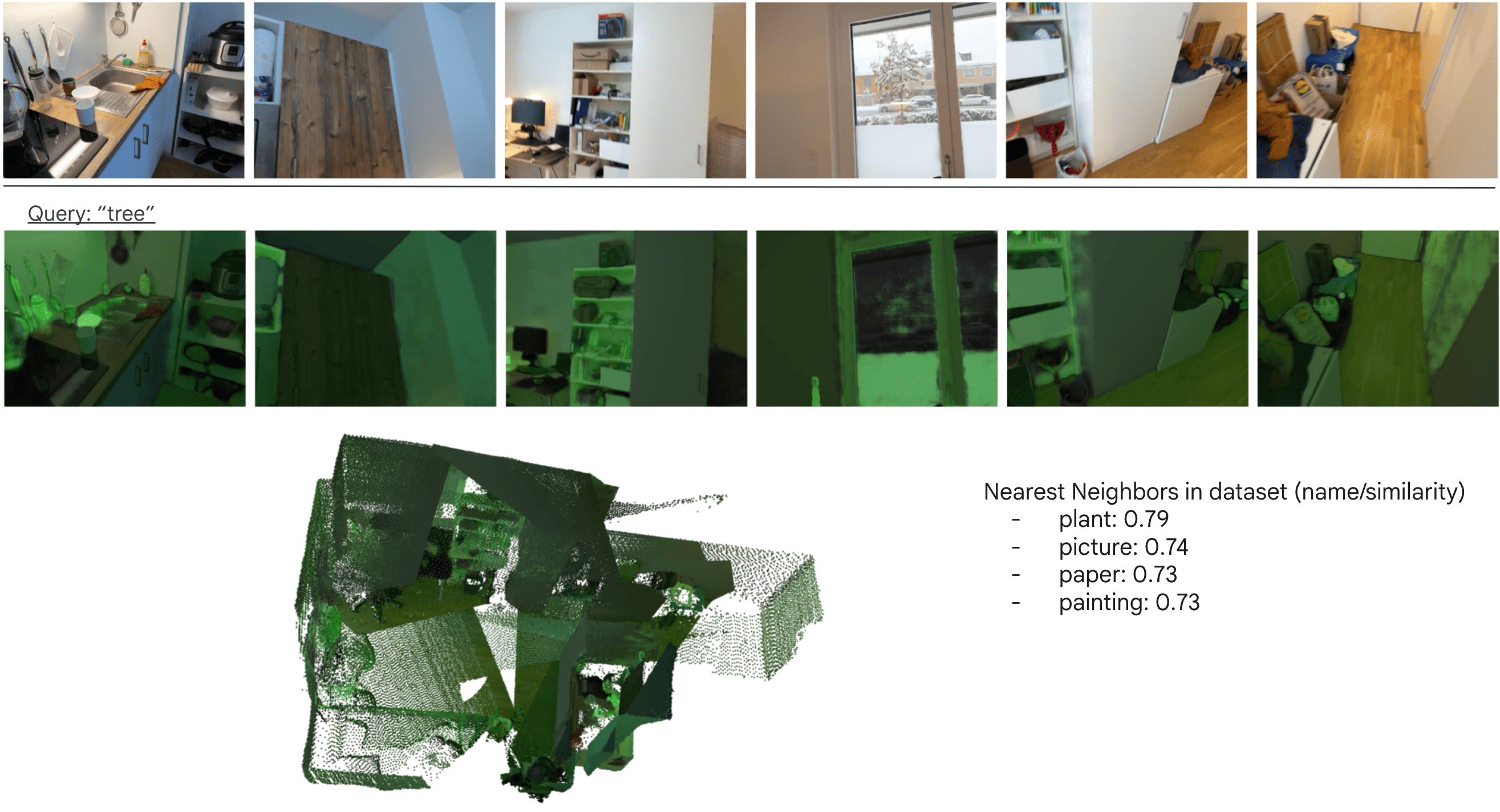}
    \vspace{-5pt}
  \end{subfigure}

  \caption{Continuation of Fig.~\ref{fig:supp:openvoc_queries}}
  \label{fig:supp:openvoc_queries2}
\end{figure}

%% file: supp_mat/S03_experiments.tex
\section{Additional experiments}
\label{sec:supp:experiments}

\noindent\textbf{Generalization.} We acknowledge that Pano3D, trained only on ScanNet++, will not generalize to drastically out-of-domain scenes like outdoors. To evaluate indoor generalization, we test on HyperSim scenes~\cite{roberts2021hypersim} as in PanSt3R using Scene Panoptic Quality (PQ) (Tab.~\ref{tab:supp:hypersim}), predicting with ScanNet++~\cite{yeshwanth2023scannet++} and then remapping to HyperSim classes. We subsample every 2 frames, and then select 25 linspaced frames for memory. Pano3D achieves a PQ of 40.4, and was trained \textit{only} on real-world ScanNet++ data, making this a real-to-synthetic zero-shot test. Ablations point out that despite a low number of keyframes, Pano3D still manages to segment the scene. We also try to substitute the text embeddings used during training (those of ScanNet++) with the HyperSim vocabulary, causing a 14 points drop in performance, due to the presence on unseen, out-of-distribution classes like 'person'.

\input{tables/supp_hypersim_inference_time}

%% file: tables/supp_hypersim_inference_time.tex
\begin{table}[h!]
    \centering
    \small %
    \begin{minipage}[t]{0.38\linewidth}
        \vspace{0pt} %
        \centering
        \setlength{\tabcolsep}{2pt}
        \caption{HyperSim PQ.}
        \label{tab:supp:hypersim}
        \resizebox{.7\linewidth}{!}{
        \begin{tabular}{lc}
            \toprule
            Method & PQ \\
            \midrule
            Pano3D & 40.4 \\
            \quad HyperSim vocab & 26.8 \\
            \quad 10 keyframes & 37.9 \\
            \quad 50 keyframes & 38.1 \\
            \bottomrule
        \end{tabular}
        }
    \end{minipage}%
    \hfill%
    \begin{minipage}[t]{0.62\linewidth}
        \vspace{0pt} %
        \centering
        \footnotesize %
        \setlength{\tabcolsep}{2.0pt} %
        \caption{Model Parameters and Inference Time.}
        \label{tab:supp:model_specs}
        \begin{tabular}{p{2.8cm}cc}
            \toprule
            Method & N. Params & Inference (s) \\
            \midrule
            Pano3D (MUSt3R~\cite{cabon2025must3r}) & 450M & 2.4 \\
            Pano3D (Pi3~\cite{wang2025pi}) & 1.1B & 10.8 \\
            \bottomrule
        \end{tabular}
    \end{minipage}
\end{table}